\documentclass[journal]{IEEEtran}
\usepackage{mathtools}
\usepackage{graphicx}
\usepackage{bm}
\usepackage{pifont}
\newcommand{\cmark}{\ding{51}}%
\newcommand{\xmark}{\ding{55}}%
\usepackage{longtable}
\usepackage{adjustbox} 
\usepackage{makecell}
\usepackage{booktabs}
\usepackage{enumitem}
\usepackage[linesnumbered,ruled]{algorithm2e}
\SetKwFor{ParFor}{parallel for}{do}{end}
\SetKwRepeat{Do}{do}{while}
\usepackage{amsmath}
\usepackage{multicol,caption}
\usepackage{color}
\usepackage{amssymb}
\usepackage{bbm}
\usepackage{subcaption} 
\usepackage{multirow}
\newenvironment{Figure}
{\par\medskip\noindent\minipage{\linewidth}}
{\endminipage\par\medskip}
\newcommand{\includegraphicsmaybe}[2][]{\IfFileExists{#2}{\includegraphics[#1]{#2}}{\includegraphics{img/dummy.pdf}}}
\usepackage{float}
\usepackage{amsmath}

\usepackage{cite}
\usepackage{CJKutf8}
\usepackage[utf8]{inputenc}
\allowdisplaybreaks[3]

\begin{document}
	
	\title{Dispersed Pixel Perturbation-based Imperceptible Backdoor Trigger for Image Classifier Models}

	\author{
		Yulong Wang, \emph{Member}, \emph{IEEE},
		Minghui Zhao,
		Shenghong Li, \emph{Member}, \emph{IEEE}, \\
		Xin Yuan, \emph{Member}, \emph{IEEE}, and
		Wei Ni, \emph{Senior Member},  \emph{IEEE}
		\thanks{Y.~Wang and M.~ Zhao are with the State Key Laboratory of Networking and Switching Technology, School of Computer Science (National Pilot Software Engineering School), Beijing University of Posts and Telecommunications, Beijing 100876, China (e-mail: \{wyl, zhaominghui\}@bupt.edu.cn).}
		\thanks{S.~Li, X.~Yuan, and W.~Ni are with the Commonwealth Science and Industrial Research Organisation (CSIRO), Marsfield, New South Wales, 2122, Australia (e-mail: \{shenghong.li, xin.yuan, wei.ni\}@data61.csiro.au).}
	}
	
	\maketitle
	\begin{abstract} Typical deep neural network (DNN) backdoor attacks are based on triggers embedded in inputs. Existing imperceptible triggers are computationally expensive or low in attack success.	In this paper, we propose a new backdoor trigger, which is easy to generate, imperceptible, and highly effective. The new trigger is a uniformly  randomly generated three-dimensional (3D) binary pattern that can be horizontally and/or vertically repeated and mirrored and superposed onto three-channel images for training a backdoored DNN model. Dispersed throughout an image, the new trigger produces weak perturbation to individual pixels, but collectively holds a strong recognizable pattern to train and activate the backdoor of the DNN. We also analytically reveal that the trigger is increasingly effective with the improving resolution of the images.	Experiments are conducted using the ResNet-18 and MLP models on the MNIST, CIFAR-10, and BTSR datasets. In terms of imperceptibility, the new trigger outperforms existing triggers, such as BadNets, Trojaned NN, and Hidden Backdoor, by over an order of magnitude. The new trigger achieves an almost 100\% attack success rate, only reduces the classification accuracy by less than 0.7\%--2.4\%, and invalidates the state-of-the-art defense techniques.
	\end{abstract}
	

	\IEEEpeerreviewmaketitle
	
	\section{Introduction}
	\label{sec: intro}
	
	In recent years, deep neural network (DNN) has been increasingly widely used thanks to its excellent performance in object detection~\cite{8627998}, voice conversion~\cite{9262021}, and autonomous driving~\cite{9000872}. DNN has even outperformed humans in some tasks, e.g., large-scale image classification~\cite{8913592, 8538003, 8673926}. However, backdoor attacks put the applications of DNN to safety- or security-critical tasks at stake. 
	Backdoored DNN was first introduced by BadNets~\cite{DBLP:journals/corr/abs-1708-06733}, where a color block was used as a trigger and added on input images to encode its features into the parameters of the backdoored DNN in the training process. Many different triggers have been proposed since, including tatoo~\cite{DBLP:conf/acsac/DoanAR20}, graffiti~\cite{DBLP:conf/cvpr/EykholtEF0RXPKS18}, and cartoon patches~\cite{DBLP:journals/corr/abs-1712-05526}.

	To counteract backdoored DNNs, a range of defense methods have been developed.
	For example, STRong Intentional Perturbation (STRIP)~\cite{DBLP:conf/acsac/GaoXW0RN19} detects triggers by assuming a distinguishable entropy distribution of the DNN model prediction between clean and poisoned images.
	Februus~\cite{DBLP:conf/acsac/DoanAR20} adopts Class Activation Map (CAM)~\cite{8237336} to 
	visualize the decision of a DNN, locate and remove the trigger, and restore the changed pixels using a pre-trained generative neural network (GAN)~\cite{9319516}.
	Spectral Signature Defense (SSD)~\cite{DBLP:conf/nips/Tran0M18} uses singular value decomposition (SVD) to obtain the dominant vector of samples in the training dataset and determine whether an input contains a trigger by comparing the correlation between the input and the vector against a pre-defined threshold. Activation Clustering (AC)~\cite{DBLP:conf/aaai/ChenCBLELMS19} treats the trigger detection as a two-class clustering problem and utilizes $K$-Means to group the inputs into two clusters based on their hidden layer outputs.
	Other defense mechanisms include Artificial Brain Stimulation (ABS)~\cite{DBLP:conf/ccs/LiuLTMAZ19}, Neuron Pruning~\cite{DBLP:conf/raid/0017DG18}, and Neural Cleanse~\cite{DBLP:conf/sp/WangYSLVZZ19}.
	These defense approaches effectively defend existing triggers that typically produce intensive perturbations to a small region of an image (e.g., sunflower trigger~\cite{DBLP:conf/acsac/DoanAR20}). 
	A strong perturbation makes the triggers distinguishable in the final and/or intermediate output of a backdoored DNN, and makes the triggers perceptible. 

	Imperceptible triggers are more threatening to DNN applications and typically studied in the contexts of DNN adversarial attack~\cite{DBLP:conf/cvpr/CohenSG20,DBLP:conf/iclr/MadryMSTV18} and DNN robustness~\cite{DBLP:conf/sp/Carlini017}. No backdoor needs to be installed in a DNN. Instead, a trigger is derived from the trained DNN to misclassify an image arbitrarily or into a target class. 
    Imperceptibility was achieved by restricting the $\ell_{\infty}$- or $\ell_2$-norm of the trigger. However, the triggers are input-specific, i.e.,
 different images (even from the same class) require different triggers, leading to high complexities in trigger generation. The triggers can also be invalidated by changes in the images. Several input-agnostic adversarial trigger generation approaches have been developed~\cite{DBLP:conf/ijcai/XiaoLZHLS18,DBLP:conf/aaai/ShafahiN0DDG20}. They exhibit much lower attack success rates (ASRs) than the input-specific imperceptible adversarial triggers. Moreover, a defender may have the knowledge of an attacker and  potentially reproduce the triggers to reinforce the DNN through so-called adversarial training~\cite{DBLP:conf/iclr/MadryMSTV18, DBLP:conf/iclr/TramerKPGBM18}. 
	
In this paper, we propose a new backdoor trigger that is easy to generate, imperceptible, and highly effective. The new trigger is a uniformly randomly generated, three-dimensional (3D) binary pattern that can be horizontally and/or vertically repeated and mirrored and then superposed onto three-channel images for training a backdoored DNN model. 
While the new trigger collectively holds a strong recognizable pattern to effectively train or activate the backdoor of a DNN model, it generates weak perturbation to individual pixels and remains imperceptible. 
The complexity of the trigger generation and image perturbation is linear to the image size, and substantially lower than those of the existing triggers. 
Extensive experiments demonstrate the superiority of the new trigger to existing methods in complexity, imperceptibility, and effectiveness. 
\begin{itemize}
    \item \textit{Low complexity:}
    	The generation of the  new trigger only involves linear operations, such as repetition, addition, and clamping (to within $[0, 255]$). The complexity is $\mathcal{O}(CWH)$ for an image with $C$ color channels and $W \times H $ pixels per channel. 
    	In contrast, the existing imperceptible triggers, e.g., {\color{black}AdvGAN~\cite{DBLP:conf/ijcai/XiaoLZHLS18},  UAT~\cite{DBLP:conf/aaai/ShafahiN0DDG20}, and Hidden Backdoor\cite{DBLP:conf/aaai/SahaSP20},} require back propagation and gradient descent operations on a neural network with dramatically higher complexities.
    	The generation of a malicious input using the new trigger is computationally efficient, since the trigger is input-agnostic and can be directly superposed onto a benign image to perturb the image. In contrast, existing approaches, such as projected gradient descent (PGD)~\cite{DBLP:conf/iclr/TramerKPGBM18} and DeepFool~\cite{DBLP:conf/cvpr/Moosavi-Dezfooli16}, produce input-dependent triggers and perturb inputs using the gradient of the attacked neural network.
    \item \textit{Imperceptibility:} 
        Compared to existing backdoor triggers typically using more visible colored blocks (e.g., \cite{DBLP:journals/corr/abs-1708-06733,DBLP:conf/acsac/DoanAR20,DBLP:conf/cvpr/EykholtEF0RXPKS18,DBLP:journals/corr/abs-1712-05526}), our trigger is imperceptible. The imperceptibility of the new trigger is analyzed both quantitatively and visually on three different datasets. We show that the proposed trigger outperforms all baseline approaches in two popular perceptual metrics, namely, Structural Similarity (SSIM) Index~\cite{1284395} and Learned Perceptual Image Patch Similarity (LPIPS)~\cite{DBLP:conf/cvpr/ZhangIESW18},
        and in the visual difference between clean and poisoned images.
    \item \textit{High {\color{black}effectiveness:}}
        The new trigger is evaluated experimentally under different datasets, neural network architectures, and latest defense strategies. We show that the trigger threatens the reliability of image classifier models with an ASR of close-to-100\%, which is 17\% higher than the existing input-agnostic imperceptible triggers, such as AdvGAN~\cite{DBLP:conf/ijcai/XiaoLZHLS18} and UAT~\cite{DBLP:conf/aaai/ShafahiN0DDG20}. The new trigger also invalidates the existing defense strategies. 
\end{itemize}

The rest of this paper is organized as follows. Section~\ref{sec: relatedwork} provides the related works. Section~\ref{sec: our_approach} presents the threat model and the new trigger. The new trigger is extensively tested against popular datasets, DNN models, and defense strategies in Section~\ref{sec: experiment}, followed by conclusions in Section~\ref{sec: conclusion}.
	
\section{Related Work}
\label{sec: relatedwork}
\subsection{Backdoor Trigger}
Recently, several backdoor triggers have been proposed, mostly visible to human eyes, such as a small patch beside a digit~\cite{DBLP:conf/sp/WangYSLVZZ19},  a pair of bright-rim eyeglasses on a face~\cite{DBLP:conf/ccs/SharifBBR16}, a bright pixel in an image~\cite{8601309}, or a cartoon watermark~\cite{DBLP:conf/acsac/GaoXW0RN19}. 
BadNets~\cite{DBLP:journals/corr/abs-1708-06733} uses a small patch (e.g., a yellow rectangle sticker on a traffic sign) as a trigger to poison images and trains a neural network model with the poisoned images.

Invisible backdoor triggers have been designed in~\cite{DBLP:conf/aaai/SahaSP20,9186317,DBLP:conf/codaspy/ZhongLSZ020,DBLP:conf/ndss/LiuMALZW018,9450029}. 
Saha \textit{et al.}~\cite{DBLP:conf/aaai/SahaSP20} introduced Hidden Backdoor (HB) attacks, which used a small image patch as a trigger. An image with invisible perturbations was produced by minimizing the difference of neuron activations in the penultimate layer between a clean image and its poisoned version with the trigger.
However, once the backdoored DNN is deployed, an attack still requires a visible trigger to activate the  embedded backdoor. Backdoor defense strategies, such as Februus~\cite{DBLP:conf/acsac/DoanAR20}, can detect and eliminate the trigger.

Li \textit{et al.}~\cite{9186317} designed two types of invisible triggers for backdoor attacks. The first type was produced by converting a static trigger (for example, a string of texts) to a binary form and then replacing the least significant bits of pixels in an image with the trigger. Only altering the least significant bits results in minor changes in color intensity, making a trigger hard to detect by human inspectors. 
However, a defender can disable the trigger by replacing the least significant bits with random values. 
The second type was to amplify the trigger-induced activations of a subset of neurons in the DNN's penultimate layer. The process also reduced the trigger's norm to below a threshold, making the trigger imperceptible. Hence, a backdoor was implanted in a small number of neurons. In the presence of the trigger, those neurons can produce significantly larger activations than the others, making them detectable for detection strategies, e.g., AC~\cite{DBLP:conf/aaai/ChenCBLELMS19}, and suppressible using neuron pruning techniques, e.g., Neural Cleanse~\cite{DBLP:conf/sp/WangYSLVZZ19}.

Zhong \textit{et al.}~\cite{DBLP:conf/codaspy/ZhongLSZ020} developed two invisible triggers. One was a static perturbation mask based on a repeated pattern. The pattern consists of an array of small sub-regions, and increases the color intensity of pixels in the sub-regions. Since this trigger is not random, it can be reverse-engineered by enumerating all possible sizes of the sub-regions and the increase of color intensity. The second trigger employed the DeepFool~\cite{DBLP:conf/cvpr/Moosavi-Dezfooli16} to obtain a universal, invisible perturbation by projecting the images of one class to the boundary between their source class and the adversary-specified target class. The adversary can poison the training data, insert the backdoor into the DNN model, and use the perturbation as a trigger. 
However, the trigger depends on the source and target classes and is computationally expensive to produce.

Trojaned NN~\cite{DBLP:conf/ndss/LiuMALZW018} and RobNet~\cite{9450029} are two recent and related designs of backdoor triggers. Trojaned NN~\cite{DBLP:conf/ndss/LiuMALZW018} selects the most connected neurons in the penultimate layer of a pre-trained neural network model, and generates a trigger that maximizes the activation of the selected neurons using gradient descent. A backdoored model is obtained by further training the model with images poisoned with the trigger. RobNet~\cite{9450029} is a variation of Trojaned NN, and conducts neuron selection and trigger generation. It supports multiple trigger locations (up to eight per image) and multiple triggers (at different locations of an image) to produce a backdoored model.
AdvGAN~\cite{DBLP:conf/ijcai/XiaoLZHLS18} and UAT~\cite{DBLP:conf/aaai/ShafahiN0DDG20} are two other recent designs of imperceptible triggers. AdvGAN adopts GANs to train a perturbation generator. For any image in the same domain as the images used to train the generator, the generator produces and adds a trigger to the image to attack a DNN model. UAT produces an input-agnostic imperceptible trigger for an image dataset and a given DNN architecture. It solves an optimization problem of universal perturbation by adopting the stochastic gradient method. AdvGAN and UAT require back propagation and gradient descent operations on a neural network, incurring high complexities.

\subsection{Defense Methods}
\label{sec: defenses}
Defense strategies have been developed to detect or disable triggers, or repair backdoored DNN models~~\cite{DBLP:conf/acsac/GaoXW0RN19, DBLP:conf/acsac/DoanAR20,DBLP:conf/nips/Tran0M18,DBLP:conf/aaai/ChenCBLELMS19,DBLP:conf/ndss/Xu0Q18,DBLP:conf/sp/WangYSLVZZ19,DBLP:conf/ccs/LiuLTMAZ19,DBLP:conf/raid/0017DG18}. 
Februus~\cite{DBLP:conf/acsac/DoanAR20} 
sanitizes inputs by removing potential trigger artifacts and keeping the information for classification tasks. The triggers are located using GradCAM~\cite{8237336}, a variant of the classic DNN visualization technique, CAM~\cite{9316944}. It was reported in~\cite{DBLP:conf/acsac/DoanAR20} that Februus reduced the ASR from 100\% to nearly 0\% for a badge, tattoo, image patch, and color block triggers on the CIFAR-10, German Traffic Sign Recognition Benchmark (GTSRB), Belgium Traffic Sign Recognition (BTSR), and VGGFace2 datasets. 

STRIP~\cite{DBLP:conf/acsac/GaoXW0RN19} is a trigger detection algorithm for vision systems. It perturbs the input to a DNN model by superimposing various image patterns, and then observes the randomness of predicted classes for the perturbed inputs. The entropy exceeding a pre-defined threshold in predicted classes indicates the presence of a backdoor trigger. STRIP archives close-to-zero false acceptance rate and false rejection rate on small black square, heart-shape frame, mosaic patch on MNIST, CIFAR-10, and GTSRB datasets. The effectiveness of STRIP depends on the existence and selection of the threshold.

SSD~\cite{DBLP:conf/nips/Tran0M18} detects triggers by first
calculating the covariance matrix of the feature representation of training samples for each class. Then, SSD calculates the correlation between the feature representation of the incoming input and the top eigenvector of the covariance matrix (i.e., the eigenvector corresponding to the largest eigenvalue). It compares the correlation with a predefined threshold to detect triggers. Let $\epsilon$ denote the ratio of poisoned data in the testing data.
SSD is effective under an $\epsilon$-spectrally separable condition, i.e.,
\begin{align}
	\text{Pr}_{X \sim \mathcal{W}} [\langle X - \mu_\mathcal{F}, v \rangle < \xi] < \epsilon \label{eqn: ssd_poison};\\
	\text{Pr}_{X \sim \mathcal{D}} [\langle X - \mu_\mathcal{F}, v \rangle > \xi] < \epsilon \label{eqn: ssd_clean},
\end{align}
where $\mathcal{W}$ and $\mathcal{D}$ are the distributions of the inner representations of the poisoned and clean samples in the hidden layer; $\mathcal{F} = \epsilon \mathcal{W} + (1-\epsilon)\mathcal{D}$ is the mixture of $\mathcal{W}$ and $\mathcal{D}$; $\mu_\mathcal{F}$ is the mean of $\mathcal{F}$;  $v$ is the top eigenvector of the covariance of $\mathcal{F}$; and $\xi$ is the threshold to distinguish clean and poisoned samples. 

AC~\cite{DBLP:conf/aaai/ChenCBLELMS19} adopts $K$-Means to cluster input images into clean and poisoned groups by inspecting the activations of the hidden layers of a DNN model. AC first retrieves the activations, then reduces their dimensions with primary component analysis (PCA)~\cite{8525438}, fast independent component analysis (Fast ICA)~\cite{8998392}, or $t$-distributed stochastic neighbor embedding (t-SNE)~\cite{9064929}. AC determines the clusters containing poisoned samples using one of four cluster analysis methods: \textit{Smaller}, \textit{Relative Size}, \textit{Distance}, and \textit{Silhouette}. The cluster with the fewest items is selected as poisoned by \textit{Smaller}. \textit{Relative Size} classifies a cluster as poisoned if the smaller one contains less data than a threshold. \textit{Distance} classifies a cluster as poisoned if its median activation is closer to the median of another class than to its own. \textit{Silhouette} analyzes the suspicion level based on size and Silhouette score~\cite{9277640}.

Neuron Pruning~\cite{DBLP:conf/raid/0017DG18} is under the premise that the average activation of neurons in the final convolutional layer of a backdoored neural network is significantly different between clean and adversarial inputs. The neurons that are dormant for clean inputs are removed to disable the backdoor. The termination condition is that the decrease of the classification accuracy on clean inputs exceeds a threshold, e.g., 4\%.

Neural Cleanse~\cite{DBLP:conf/sp/WangYSLVZZ19} is a trigger detection and mitigation method for DNN backdoor attacks. It detects and reverse-engineers a trigger by finding the minimal trigger required to misclassify all samples from other labels to the target label. The trigger size is measured by the number of pixels replaced. The mitigation techniques include \textit{input filtering}, \textit{neuron pruning}, and \textit{unlearning}. \textit{Input filtering} discards inputs with potential triggers. \textit{Neuron pruning} removes backdoor-related neurons identified by the reverse-engineered trigger in the penultimate layer of the DNN model. \textit{Unlearning} trains the backdoored DNN to forget the trigger by using the reverse-engineered trigger and correct labels.	

Spatial Smoothing~\cite{DBLP:conf/ndss/Xu0Q18} is a class of widely used techniques in image processing for suppressing image noise. Local smoothing methods make use of nearby pixels to smooth each pixel. By selecting different weighting mechanisms for neighboring pixels, a local smoothing method can be Gaussian, mean, or median smoothing. The median filter runs a sliding window over each pixel, where the center pixel is replaced by the median value of the neighboring pixels within the window. 
Spatial Smoothing is effective in defending malicious inputs having invisible triggers, such as adversarial examples generated by PGD~\cite{DBLP:conf/iclr/MadryMSTV18}.

Last but not least, adaptive attacks can minimize the difference of neuron activations within each layer or reduce the difference of neuron activations between benign and malicious inputs (e.g., by minimum or min-max criteria) during data poisoning. ABS~\cite{DBLP:conf/ccs/LiuLTMAZ19} provides an effective means to defend adaptive attacks by defining an adaptive loss function and minimizing it together with the classification loss function.

As will be shown in Section~\ref{sec: resist}, none of the above state-of-the-art techniques is effective in detecting or disabling the new backdoor trigger discovered in this paper.

\section{New Backdoor Trigger to Image Classification Neural Networks}
\label{sec: our_approach}
In this section, we first describe the threat model of the new backdoor trigger. Then, we describe the trigger generation, followed by implementation considerations.

\subsection{Threat Model}
\label{sec: threatmodel}
We adopt a threat model similar to the one used in~\cite{9450029}, where a cloud service provider delivers Machine-Learning-as-a-Service (MLaaS) Platform. DNN training usually requires significant computing resources, e.g., GPU clusters, and domain knowledge of the DNN design and hyperparameter configuration. Commercially available MLaaS platforms, e.g., AWS Machine Learning, Google Cloud Machine Learning, and Microsoft Azure ML Studio, are widely accessible for DNN users to outsource DNN model training. In this case, the cloud service provider can be adversarial (because of a rogue employee or a compromised server). A user outsourcing the DNN training to the cloud provider is the defender. 
The DNN user evaluates the prediction performance of the received DNN model with clean datasets (e.g., images). The user may also try to detect the backdoors in the DNN model, and disable backdoor triggers using the state-of-the-art algorithms.
	
The goal of an adversary performing a backdoored DNN attack can be formulated as the following Maximum Likelihood Estimation (MLE) problem:
	\begin{align}
		\max_{\mathbf{T},\mathbf{\theta}} \prod_{i=1}^{N} \phi_{\mathbf{\theta}}(c_{\mathbf{X}^{(i)}}|\mathbf{X}^{(i)})
		\phi_{\mathbf{\theta}}(c_t|\mathcal{T}(\mathbf{X}^{(i)} + \mathbf{T}))^{p_i},
		\label{eqn:trigger_learn}
	\end{align}
where 
\begin{itemize}
\item $\mathbf{X}^{(i)} \in [0,255]^{L_H \times L_V \times N_C}$ ($i = 1, \dots, N$) is the $i$-th testing image with the size of $L_H \times L_V$ and $N_C$ color channels; 
	
\item $\mathbf{T} \in [-255,255]^{L_H \times L_V \times N_C}$ is the backdoor trigger; 

\item $p_i \in \{0,1\}$ indicates whether $\mathbf{X}^{(i)}$ is poisoned by trigger $\mathbf{T}$ for backdoor learning or not ($p_i = 1$ if $\mathbf{X}^{(i)}$ is poisoned; or $p_i = 0$, otherwise);	

\item $c_{\mathbf{X}^{(i)}}$ is the ground-truth class of $\mathbf{X}^{(i)}$;
	
\item $c_t$ is the target class specified by the attacker; 

\item $\mathbf{\theta}$ is the set of the model parameters of the DNN, i.e., the weights of connections and the biases of neurons; 

\item $\phi_{\mathbf{\theta}}(c|\mathbf{X})$ is the trained DNN which outputs the  probability of the input $\mathbf{X}$ belonging to class $c \in \{c_t,c_{\mathbf{X}^{(i)}},\forall i\}$;

\item $\mathcal{T}(\mathbf{X})$ is a truncation function to ensure that each $N_C$-channel pixel of a poisoned image $\mathbf{X}$ is within $[0, 255]^{N_C}$.

\end{itemize}
The adversary trains the DNN model to misclassify any poisoned input (i.e., images embedded with the trigger $\mathbf{T}$) to the target class specified by the adversary. For clean inputs (e.g., images without the trigger $\mathbf{T}$), the adversary wishes the DNN model to provide a satisfactory classification accuracy.

The joint optimization can be readily decoupled into the separate design and optimization of the trigger $\mathbf{T}$ and the model parameter~$\theta$. Specifically, $\mathbf{T}$ can be viewed as an additional feature of the images, and is used to classify the images into a target class specified by the attacker, as done in existing backdoored models, e.g., BadNets~\cite{DBLP:journals/corr/abs-1708-06733}. According to the Universal Approximation Theorem~\cite{DBLP:journals/natmi/LuJPZK21}, a neural network can approximate any continuous function at any given precision requirement. In this sense, given any trigger $\mathbf{T}$, the model parameter $\theta$ can be trained to achieve any pre-specified classification accuracy of clean images and any pre-specified misclassification rate of poisoned images. For this reason, the attacker could design $\mathbf{T}$, prior to the training of $\theta$ based on the trigger~$\mathbf{T}$, as done in 
BadNets~\cite{DBLP:journals/corr/abs-1708-06733}.

\subsection{The New Backdoor Trigger}
\label{sec: trigger_desc}
We discover a new backdoor trigger $\mathbf{T}$, which is a randomly generated binary 3D matrix superposed to the pixels of the input image. A DNN trained with poisoned images, $\mathbf{X'}$, can detect the presence of $\mathbf{T}$ and, therefore, contain a backdoor that can be exploited in a later stage. By carefully configuring its magnitude $m$, the backdoor trigger $\mathbf{T}$ can be imperceptible to human eyes.  $\mathbf{T}$ is also input-agnostic, since it is generated independently of input images.
    
\subsubsection{Generation of the New Trigger}\label{sec: trigger generation}
\begin{figure}
    \centering
    \includegraphics[height=3in]{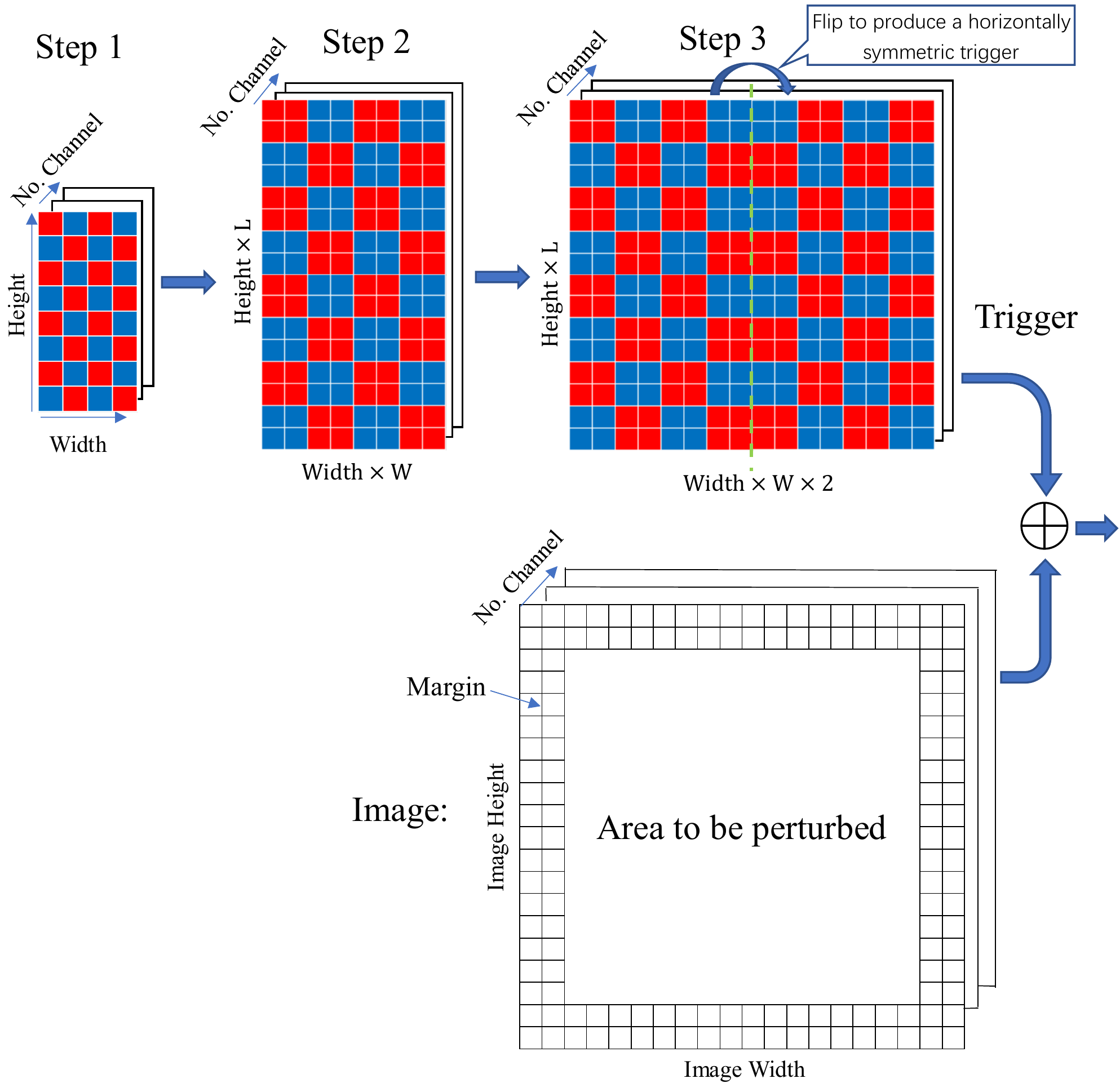}
    \caption{{\color{black}The generation process of the proposed trigger, where a red box corresponds to a ``$+m$'' and a blue box corresponds to a ``$-m$''.}}
    \label{fig:my_trigger}
\end{figure}

As illustrated in Fig.~\ref{fig:my_trigger}, the new trigger is generated in three steps.
\begin{enumerate}[label=Step \arabic*:, leftmargin=*, widest=iii]
    \item  Produce a 3D random matrix with the dimension of $T_H \times T_V \times N_C$, where each layer comprising the first two dimensions of the matrix corresponds to one channel of the images to be poisoned. 
    The third dimension indicates different channels of the images.	The elements of the 3D matrix follow the i.i.d. binary distribution with amplitude $m$ (i.e.,  ``$\pm m$'') and are generated using the Cryptographically Secure Pseudo-Random Number Generator (CSPRNG). 
    
    \item  Extend each layer of the 3D matrix using repetition, i.e., repeating every element horizontally and vertically multiple times on each layer. 
    
    \item Mirror the 3D matrix horizontally (and/or vertically) on each layer to produce a horizontally (and/or vertically) symmetric matrix used as the new trigger. 
\end{enumerate}
When poisoning an image, each page of the trigger is superposed (i.e., added) onto the middle of the corresponding channel of the image with margins reserved unperturbed to bypass Februus trigger removal~\cite{DBLP:conf/acsac/DoanAR20}. The resulting magnitude of each pixel is truncated to be within the valid range of $[0,255]^{N_C}$. The purposes of the repetition in Step 2 and the symmetric extension in Step 3 are to get around the typical image blurring method, such as Spatial Smoothing~\cite{DBLP:conf/ndss/Xu0Q18}, and image transformation, such as flipping. 

The numbers of horizontal and vertical repetitions per pixel in Step 2, denoted by $R_H$ and $R_V$, and the width of the unperturbed margin, denoted by $M_G$, are hyperparameters and can be adjusted, given the size of the 3D random matrix generated in Step 1, i.e., $T_H\times T_V\times N_C$, and the size of the images to be poisoned, i.e., $L_H\times L_V\times N_C$. Apparently, $L_H=2(R_HT_H+M_G)$ and $L_V=R_VT_V+2M_G$ in Fig.~\ref{fig:my_trigger}.

A backdoor trigger is a secret withheld by an attacker, and  
can be constructed from a random sequence generated by a pseudo-random number generator. Most statistical Pseudo-Random Number Generators (PRNGs) are based on recurrences and can be recognizable by assessing output streams~\cite{SHEMA2012209}. It is possible to predict the future output of PRNGs~\cite{Lovric2011} based on past observations. In contrast, a CSPRNG avoids detectable regularities and can withstand cryptanalysis conducted by a defender with full knowledge of the CSPRNG used. Given a sequence of pseudo-random bits generated by a randomly initialized CSPRNG, it is impossible to predict the next bit with a probability greater than 1/2 using a probabilistic polynomial-time algorithm.

The proposed trigger is much easier to generate than the existing imperceptible triggers. The generation of the new trigger only involves linear operations, such as repetition, addition, and clamping (to within $[0, 255]$). It incurs the computational complexity of $\mathcal{O}(CWH)$ for an image with $C$ color channels and $W \times H $ pixels per channel. In contrast, the existing imperceptible triggers,  such as AdvGAN~\cite{DBLP:conf/ijcai/XiaoLZHLS18}, UAT~\cite{DBLP:conf/aaai/ShafahiN0DDG20}, and Hidden Backdoor~\cite{DBLP:conf/aaai/SahaSP20}, require back propagation and gradient descent operations on a neural network. Their computational complexities are significantly higher than 
$\mathcal{O}(CWH)$.

\subsubsection{Selection of Trigger's Magnitude}
We analyze the required magnitude of the proposed backdoor trigger for specific ASRs. Our analysis based on a Perceptron model suggests that the new trigger is not only feasible, but effective in creating a backdoor and poisoning the model.

Let $\mathbf{P}$ denote a 3D random matrix generated using the CSPRNG. For illustration convenience, we assume that the size of $\mathbf{P}$, i.e., $T_H\times T_V\times N_C$, is the same as the size of an input image $\mathbf{X}$, i.e., $L_H\times L_V\times N_C$. Then, the new trigger is $\mathbf{T} = m\mathbf{P}$, where $m\in \mathbb{R}^+$ is the magnitude of the trigger. The size of the trigger is $M=T_H T_V N_C$.

\begin{figure}
		\centering
		\includegraphics[width=5cm]{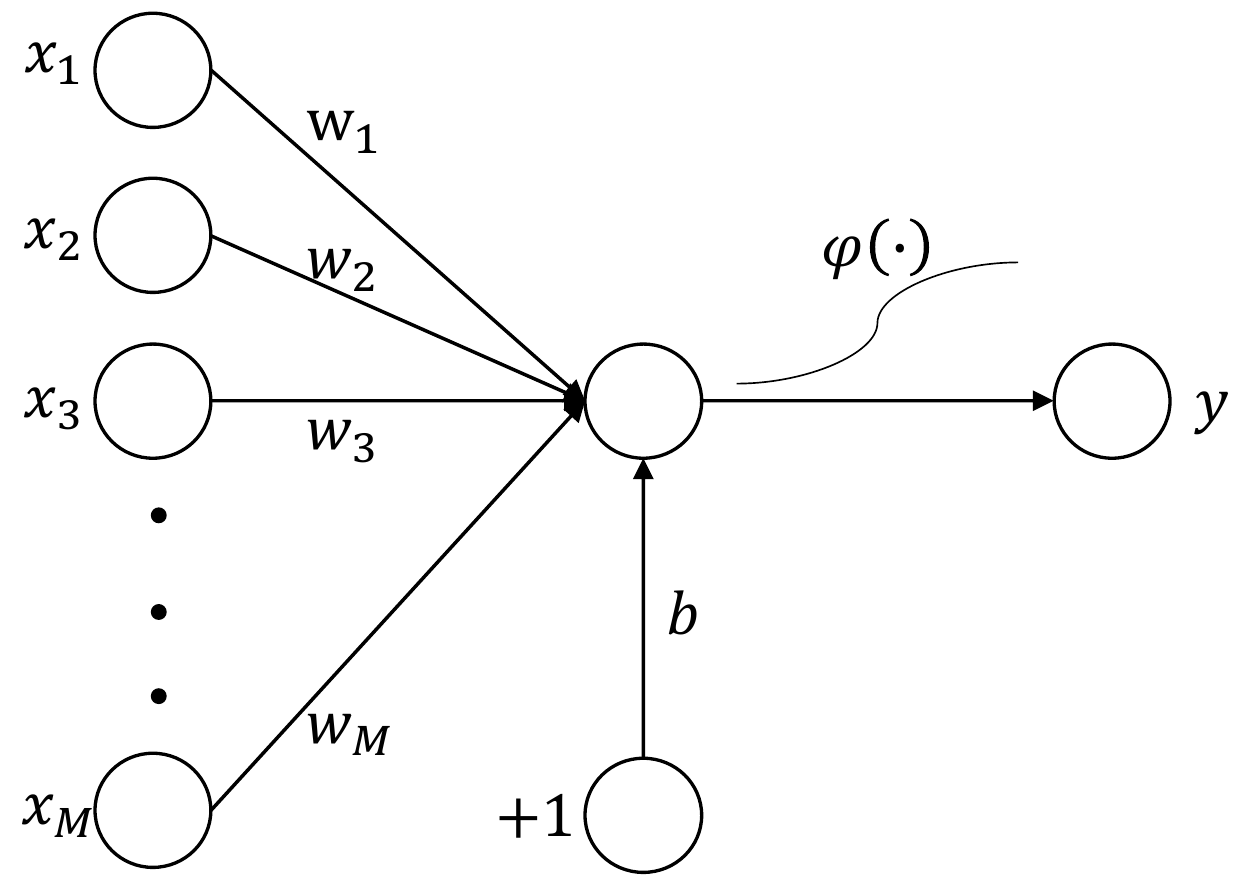}
		\caption{A Perceptron is trained to recognize the new trigger, where the activation function $\varphi(\cdot)$ maps positive and negative inputs to the presence and absence of the trigger, respectively.}
		\label{fig: perceptron}
	\end{figure}

Considering the Perceptron model in Fig.~\ref{fig: perceptron}, which takes vectorized images as the input.  The magnitude of the trigger, $m$, is obtained by solving the following problem:
	\begin{align}
		&\min_{m,\mathbf{w} }m  \nonumber \\ 
		s.t.~~~
    	 &\text{Pr}\left\{\mathcal{T}(\mathbf{x} + m\mathbf{p})^{\top} \mathbf{w} + b > 0\right\} > \eta
    	 , \label{eqn:perceptron}\\
		 &\text{Pr}\left\{\mathbf{x}^{\top} \mathbf{w} + b < 0\right\} > \eta, \nonumber
	\end{align}	
	where $\mathbf{p}$
	and $\mathbf{x}$
	are the vectorizations of $\mathbf{P}$ and $\mathbf{X}$;
	$\mathcal{T}(\mathbf{x} + m\mathbf{p})$ is the poisoned version of $\mathbf{x}$, see~\eqref{eqn:trigger_learn}; and $\eta$ is the ASR.

Since the elements in $\mathbf{p}$ are i.i.d., $\mathbb{E}[\mathbf{x}^T\mathbf{p}]=0$ and $\mathbf{p}^T\mathbf{p}=M$. By vectorizing $\mathbf{T}$ to $\mathbf{t}$, we set $\mathbf{w}=\mathbf{t}=m\mathbf{p}$ and $b=-\frac{m^2M}{2}$ in (\ref{eqn:perceptron}). Then,
	\begin{align}\label{eqn: weighting}
		(\mathbf{x} + \mathbf{t})^{\top}  \mathbf{t}  = \sum_{
		i
		}(x_{i}  t_{i}) + m^2 M, 
	\end{align}
where $x_{i} $ and $t_{i} $ are the $i$-th elements of $\mathbf{x}$ and $\mathbf{t}$, respectively. 
	
Assume that the pixels are independent in the image $\mathbf{x}$. Then, $\mathbf{x}^{\top}  \mathbf{t}$ obeys the Gaussian distribution according to the Law of Large Numbers. Also assume that ${x}_{i} \thicksim  \mathcal{U}[0,255]$. Then, $\mathbb{E}[{x}_{i}] = \frac{255}{2}$ and  $\mathbb{D}[{x}_{i}]= 255^2/12$. Since $\mathbb{E}[{t}_{i}] = 0$ and $\mathbf{t}$ is independent of $\mathbf{x}$, we have $\mathbb{E}[\mathbf{x}^{\top} \mathbf{t}] = 0$. Then, the variance of $\mathbf{x}^{\top} \mathbf{t}$ is given by 
	\begin{align}
		\mathbb{D}[\mathbf{x}^{\top} \mathbf{t}] 
		 = M m^2  C, 
	\end{align}
	where $C = \mathbb{E}[{x}_{i}]^2 + \mathbb{D}[{x}_{i}]$. As a result,
	\begin{align}
		\mathbb{E}[(\mathbf{x} + \mathbf{t})^{\top}  \mathbf{t}] = M m^2; \,\,
		\mathbb{D}[(\mathbf{x} + \mathbf{t})^{\top}  \mathbf{t}] = \mathbb{D}[\mathbf{x}^{\top}  \mathbf{t}] . 
	\end{align}
	
The rate of the Perceptron recognizing the backdoor trigger in Fig.~\ref{fig: perceptron} (i.e., the ASR) is given by
	\begin{align}
	\nonumber
	\text{Pr}\left\{\mathcal{T}(\mathbf{x} + m\mathbf{p})^{\top} \mathbf{w} + b > 0\right\} = 
	\text{Pr}\left\{\mathbf{x}^{\top} \mathbf{w} + b < 0\right\} =  \\
	1 - Q\left(\frac{\mathbb{E}[(\mathbf{x} + \mathbf{t})^{\top}  \mathbf{t}]}{2\sqrt{\mathbb{D}[\mathbf{x}^{\top} \mathbf{t}]}}\right)
	= 1- Q\left(\frac{m}{2}\sqrt{\frac{M}{C}}\right), 
	\label{eqn: lowerbound1}
	\end{align}
where $Q(\cdot)$ is the tail distribution function of the Normal distribution. A large value of $m$ increases the ASR (at an increased risk of the trigger being perceptible to human eyes). Given the ASR $\eta$, the magnitude $m$ satisfies
	\begin{align}
	\label{eqn:calc_m}
	    m \ge 2 Q^{-1}(\eta)\sqrt{\frac{C}{M}},
	\end{align}
where $Q^{-1}(\cdot)$ is the inverse function of $Q(\cdot)$. 
The right-hand side of \eqref{eqn:calc_m} provides a lower bound for $m$, since  $\mathbf{w}$ and $b$ in \eqref{eqn:perceptron} are designed only for trigger recognition.
The lower bound confirms that a higher-resolution image is more susceptible to backdoor attacks since the image can accommodate a longer and less visible trigger (with a larger $M$ and a smaller $m$).
To this end, the new trigger may be less effective when small-sized, black and white images are perturbed. This is because the trigger can be more visible on black and white images, especially when the images are small and the amplitude of the trigger needs to be large to be effective.

As discussed in Section~\ref{sec: trigger generation}, repeating each randomly generated $\pm m$ element horizontally and vertically (over $R_H$ and $R_V$ pixels, respectively) and then mirroring horizontally (and/or vertically) on each layer of the 3D matrix to produce a horizontally (and/or vertically) symmetric 3D trigger, are recommended to get around Februus trigger removal~\cite{DBLP:conf/acsac/DoanAR20} and image transformation-based trigger disabling~\cite{Shorten2019}. 
In the experiments presented in Section~\ref{sec: experiment}, we repeat every random element horizontally and vertically and then mirror it horizontally to produce a horizontally symmetric trigger. While the trigger size is $M$, the number of randomly generated elements in the trigger is $M/(2R_HR_V)$ (since each element is replicated for $2R_HR_V$ times, one per pixel). The magnitude assigned to $2R_HR_V$ pixels is $m'\geq 2 Q^{-1}(\eta)\sqrt{\frac{C}{M/2R_HR_V}}$. The magnitude needs to be distributed evenly among the $2R_HR_V$ pixels. The per-pixel magnitude $m$ yields $m'=R_HR_Vm\geq 2 Q^{-1}(\eta)\sqrt{\frac{C}{M/2R_HR_V}}$, or
\begin{equation}\label{eq: analysis with repetition and flipping}
    m\geq \frac{2 Q^{-1}(\eta)}{\sqrt{2R_HR_V}}\sqrt{\frac{C}{M}}. 
\end{equation}

\subsubsection{Implementation Consideration}
Defenders may apply image transformation techniques (e.g., random cropping, flipping, or rotation) to disable a trigger. To bypass these defenses, the trigger can be generated to be horizontally and vertically symmetric. Data augmentation can also be applied when training the backdoored DNN, by generating  randomly cropped and rotated versions of poisoned images.
Defenders may detect triggers by observing the contribution of pixels to classification through DNN visualization techniques, such as CAM~\cite{8237336}. A smaller-sized trigger can be generated only to perturb the inner part of an image. The margins remain intact to escape the inspection of activation in the margins. 

According to the moments of the distribution of input images, the images are often normalized in DNN training. To prevent the normalization of input images from destroying the trigger, the attacker can unnormalize the trigger $\mathbf{T}$ before applying it to an image. The unnormalization can be given by $\tilde{\mathbf{T}} =  \mathbf{T} \times \mathbf{\sigma} + \mathbf{\mu} $, where $\tilde{\mathbf{T}}$ is the unnormalized trigger; $\mathbf{\mu}$ and $\mathbf{\sigma}$ are the mean and standard deviation of the input images, respectively. Since $\mathbf{\mu}$ and $\mathbf{\sigma}$ are publicly known, the unnormalization can be readily accomplished by the adversary.

\section{Experiment Results}
\label{sec: experiment}
In this section, we experimentally evaluate the proposed backdoor trigger in terms of the attack capability and its resilience toward popular defense strategies.

\subsubsection{DNN Model}
We evaluate the threat of the new backdoor trigger by considering a customized Multilayer Perceptron (MLP) and a Convolutional Neural Network (CNN).
\begin{itemize}
	\item \textbf{MLP}: We consider a 7-layer MLP made up of three fully-connected layers, two Leaky ReLU activation layers with a negative slope of 0.2, and two Dropout Layers with the dropout probability of~0.2; see~\cite{9186307}; 
	
	\item \textbf{CNN}: We choose the 18-layer ResNet, where, besides the heading convolution layer and the fully-connected layer at the end of the CNN, the remaining sixteen convolutional layers in the middle are grouped into eight pairs. In each pair, a skip connection adds the input of the first convolutional layer to the output of the second convolutional layer; see~\cite{8984747}.
\end{itemize}

\subsubsection{Dataset}
The following public datasets produce poisoned data {\color{black}for training and testing} backdoored DNNs. In each training dataset, 5\% of images are poisoned with the new backdoor trigger. We train all three datasets with 100 epochs using a Stochastic Gradient Decent optimizer with the learning rate of 0.1, momentum 0.9, and weight decay $5\times 10^{-4}$.
\begin{itemize}
	\item \textbf{MNIST~\cite{6296535}}: MNIST is a handwritten digital dataset, consisting of 60,000 training and 10,000 testing samples. Each sample is a $28 \times 28$ gray-scale image. Any class in the dataset can be chosen as the target class. We choose \textit{five} as the target class. The trigger used is horizontally symmetric with a margin of $4$ and $4$ horizontal and vertical repetitions;
	
	\item \textbf{CIFAR-10~\cite{9031131}}: CIFAR-10 is a low-resolution natural image dataset with ten classes. Each sample is a $32 \times 32 \times 3$ color image. The numbers of training and testing data samples are 50,000 and 10,000, respectively. The trigger is horizontally symmetric with a margin of $4$ and $4$ horizontal and vertical repetitions. We choose \textit{dog} as the target class;
	
	\item \textbf{BTSR~\cite{9208309}}: BTSR contains 62 classes of high-resolution images typically resized to $224 \times 224 \times 3$. 
	We choose \textit{class 5} at random as the target. 
	BTSR contains only 4,570 training samples and 2,528 testing samples. The trigger is horizontally symmetric with a margin of $20$ and 14 horizontal and vertical repetitions. 
\end{itemize}
Table~\ref{tab:magnitude} provides the analytical lower bound of the trigger magnitude $m$ for the MNIST, CIFAR-10, and BTSR datasets based on \eqref{eq: analysis with repetition and flipping}, and the default $m$ values used in the experiments.

\begin{table}
\caption{{\color{black}Analytic lower bound for the trigger magnitude $m$ under $\eta=99.9\%$, and the $m$ values taken in the experiments. }}
\label{tab:magnitude}
\centering
 \renewcommand{\arraystretch}{1.2}
 \renewcommand\tabcolsep{2.5pt}
\begin{tabular}{c|c|c|c|c|c}
		    \hline
			Dataset&Repetition &Symmetry &$M$  & Analytical $m$ & Selected $m$ \\
			\hline   
			MNIST&4&Horizontal &576&5.94&10\\
			CIFAR-10&4&Horizontal
			&2325& 2.94&4\\
			BTSR&14& Horizontal
			&124848&0.12&3\\
			\hline
\end{tabular}
\end{table}

\subsubsection{Evaluation Metric}
We evaluate the following metrics for the attack performance of the new backdoor trigger:
\begin{itemize}
	\item \textbf{Functionality~\cite{9186317}}: The average classification accuracy of a DNN when tested only using clean images;
	
	\item \textbf{Functionality Loss (Func. Loss)}: The difference in Functionality between a benign DNN (trained using clean images) and a backdoored DNN;
	
	\item \textbf{Attack Success Rate (ASR)}: The ratio of the poisoned images classified correctly to the target class to the total number of images in the testing dataset. The poisoned images are generated by embedding {\color{black}a} 
	trigger to the clean images in the testing dataset;
	
	\item \textbf{Balanced Accuracy (bACC)~\cite{5597285}}: This is the arithmetic mean of the true positive rate (TPR) and the true negative rate (TNR) of poisoned data classification, as given by
\begin{align}
			\text{bACC} &= (\text{TPR}+\text{TNR})/2,
\end{align}
where $\text{TPR} = \frac{\text{\# identified poisoned samples}}{\text{\# all poisoned samples}}$ and $\text{TNR} = \frac{\text{\# identified clean samples}}{\text{\# all clean samples}}$. 
This metric quantifies the trigger detection accuracy of the considered defense algorithms.
\end{itemize}

We adopt two perceptual metrics to quantify the imperceptibility of a trigger in a poisoned image: 
\begin{itemize}
    \item \textbf{Structural Similarity (SSIM) Index}~\cite{1284395}, an extensively adopted metric for measuring the structural similarity of two images; 
    \item  \textbf{Learned Perceptual Image Patch Similarity (LPIPS)}~\cite{DBLP:conf/cvpr/ZhangIESW18}, a metric designed specifically to quantify the invisibility of triggers in poisoned images.
\end{itemize}

\subsection{Attack Success Rate}
{\color{black}The attacking performance of the new trigger is shown in Table~\ref{tab: asr_duplicate}. Since the trigger is random, we repeat these experiments with five independently randomly generated triggers on the three datasets. In general, the ASRs are close-to-100\% under all the considered models and datasets. The backdoor trigger can be easily recognized by the models, and exploited by the adversary. At the same time, the Functionality Loss is low, only around 2\%. 
The image classification capability of the poisoned DNN models is not compromised by the backdoor trigger, hence making the trigger hard to notice. As also shown in Table~\ref{tab: asr_duplicate}, any trigger with the same magnitude and margin can achieve nearly the same performance (including the ASR and the Functionality Loss) in the test stage.}

	\begin{table}
    \caption{The attack capability of the new trigger and its impact on the Functionality Loss of the considered DNN models.}
    \label{tab: asr_duplicate}
    \centering
    \begin{tabular}{c|c|c|c}
    		    \hline
    			Model&Dataset&ASR  & Functionality Loss \\
    			\hline   
    			&&	96.80\% &0.70\% \\
    			&&	98.11\% &1.14\% \\
    			MLP&MNIST&	95.92\% &0.81\% \\
    			&&	95.88\% &0.78\% \\
    			&&	96.67\% &0.89\% \\
    			\hline
    			&& 95.30\% &2.40\% \\
    			&& 95.95\% &0.49\% \\
    			ResNet-18&CIFAR-10& 97.63\% &0.98\% \\
    			&& 96.30\% &0.78\% \\
    			&& 96.71\% &0.97\% \\
    			\hline
    			&& 98.70\% &2.10\% \\
    			&& 98.43\% &2.37\% \\
    			ResNet-18&BTSR& 98.46\% &1.73\% \\
    			&& 98.69\% &2.40\% \\
    			&& 97.25\% &1.38\% \\
    			\hline
    \end{tabular}
    \end{table}

\subsection{Imperceptibility}
Table~\ref{tbl:mnist_visibility} evaluates quantitatively and qualitatively the invisibility of the proposed trigger on the MNIST dataset. It shows that the perturbation magnitude $m=10$ can provide reasonable imperceptibility. The difference between a clean image and its poisoned version is unnoticeable in the residual maps, as also corroborated quantitatively with the SSIM close to one and the LPIPS close to zero. 
    \begin{table}
    \caption{{\color{black}The qualitative and quantitative evaluation of the imperceptibility of the new trigger on the MNIST dataset, where the perturbation magnitude $m$ is set to 10.}} 
    \label{tbl:mnist_visibility} 
\centering
    \begin{tabular}{m{4em}m{4em}m{6em}m{4em}m{3em}}
    \hline 
     Origin & Poisoned& Residual Map& SSIM& LPIPS \\ 
    \hline 
		\adjustbox{center,valign=m,margin=1}{\includegraphics[height=0.2in]{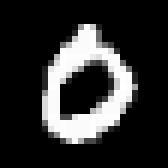}} & \adjustbox{center,valign=m,margin=1}{\includegraphics[height=0.2in]{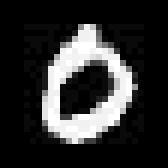}} & \adjustbox{center,valign=m,margin=1}{\includegraphics[height=0.2in]{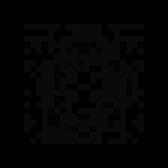}}& 
		0.993159 & 
		0.004201\\
		\hline
		\adjustbox{center,valign=m,margin=1}{\includegraphics[height=0.2in]{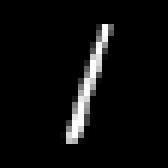}} & \adjustbox{center,valign=m,margin=1}{\includegraphics[height=0.2in]{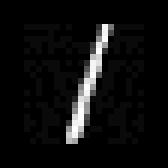}} & \adjustbox{center,valign=m,margin=1}{\includegraphics[height=0.2in]{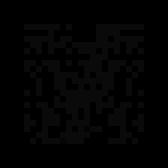}}& 
		0.993159 & 
		0.004201\\
		\hline
		\adjustbox{center,valign=m,margin=1}{\includegraphics[height=0.2in]{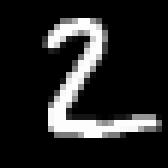}} & \adjustbox{center,valign=m,margin=1}{\includegraphics[height=0.2in]{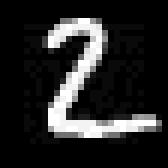}} & \adjustbox{center,valign=m,margin=1}{\includegraphics[height=0.2in]{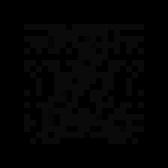}}& 
		0.993159 & 
		0.004201\\
		\hline
		\adjustbox{center,valign=m,margin=1}{\includegraphics[height=0.2in]{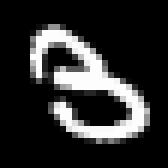}} & \adjustbox{center,valign=m,margin=1}{\includegraphics[height=0.2in]{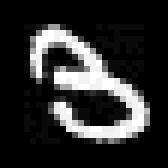}} & \adjustbox{center,valign=m,margin=1}{\includegraphics[height=0.2in]{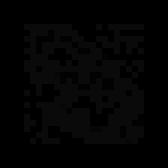}}& 
		0.988416 & 
		0.045952\\
		\hline
		\adjustbox{center,valign=m,margin=1}{\includegraphics[height=0.2in]{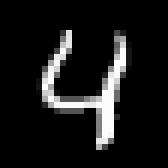}} & \adjustbox{center,valign=m,margin=1}{\includegraphics[height=0.2in]{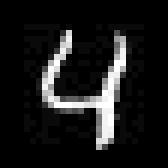}} & \adjustbox{center,valign=m,margin=1}{\includegraphics[height=0.2in]{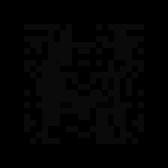}}& 
		0.946696 & 
		0.175602 \\
		\hline
		\adjustbox{center,valign=m,margin=1}{\includegraphics[height=0.2in]{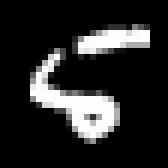}} & \adjustbox{center,valign=m,margin=1}{\includegraphics[height=0.2in]{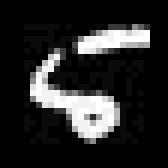}} & \adjustbox{center,valign=m,margin=1}{\includegraphics[height=0.2in]{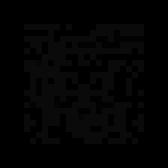}}& 
		0.993159 & 
		0.004201\\
		\hline
		\adjustbox{center,valign=m,margin=1}{\includegraphics[height=0.2in]{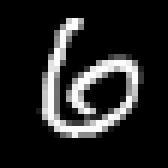}} & \adjustbox{center,valign=m,margin=1}{\includegraphics[height=0.2in]{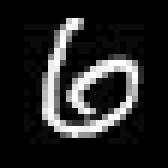}} & \adjustbox{center,valign=m,margin=1}{\includegraphics[height=0.2in]{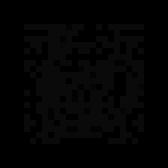}}& 
		0.993159 & 
		0.004201\\
		\hline
		\adjustbox{center,valign=m,margin=1}{\includegraphics[height=0.2in]{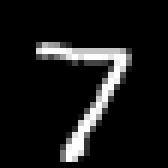}} & \adjustbox{center,valign=m,margin=1}{\includegraphics[height=0.2in]{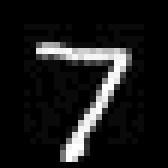}} & \adjustbox{center,valign=m,margin=1}{\includegraphics[height=0.2in]{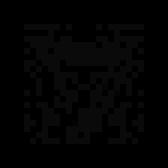}}& 
		0.985550 & 
		0.014404\\
		\hline
		\adjustbox{center,valign=m,margin=1}{\includegraphics[height=0.2in]{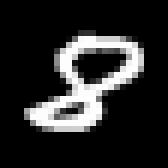}} & \adjustbox{center,valign=m,margin=1}{\includegraphics[height=0.2in]{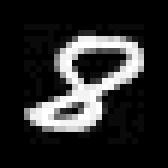}} & \adjustbox{center,valign=m,margin=1}{\includegraphics[height=0.2in]{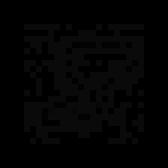}}& 
		0.985550 & 
		0.014404\\
		\hline
		\adjustbox{center,valign=m,margin=1}{\includegraphics[height=0.2in]{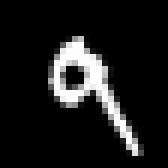}} & \adjustbox{center,valign=m,margin=1}{\includegraphics[height=0.2in]{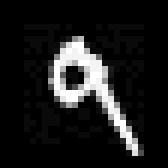}} & \adjustbox{center,valign=m,margin=1}{\includegraphics[height=0.2in]{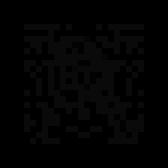}}& 
		0.993159 & 
		0.004201\\
		\hline
\end{tabular}
    \end{table}
    Table~\ref{tab:CIFAR-10_diff} evaluates the invisibility of the trigger on the CIFAR-10 dataset. Without loss of generality, we select at random a clean image from each of the ten classes in the CIFAR-10 dataset. 
    We see that the difference between a clean image and its poisoned version is unnoticeable in the residual maps, and the SSIM is close to one and the LPIPS is close to zero.

    \begin{table}
    \caption{An illustration of the visibility of the proposed trigger on the CIFAR-10 dataset.} \label{tab:CIFAR-10_diff}
    \centering
    \begin{tabular}{m{3em}m{4em}m{4em}m{4em}m{3em}m{3em}}
    \hline 
     Class & Origin & Poisoned& Residual& SSIM& LPIPS  \\ 
    \hline 
		airplane&
		\adjustbox{center,valign=m,margin=1}{\includegraphics[height=0.5in]{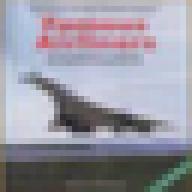}} & \adjustbox{center,valign=m,margin=1}{\includegraphics[height=0.5in]{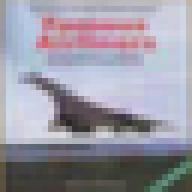}} & \adjustbox{center,valign=m,margin=1}{\includegraphics[height=0.5in]{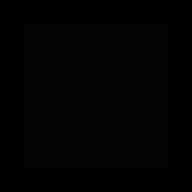}}& 
		0.979512&0.003648\\
		\hline
		car&\adjustbox{center,valign=m,margin=1}{\includegraphics[height=0.5in]{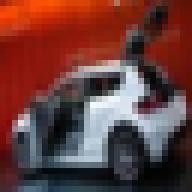}} & \adjustbox{center,valign=m,margin=1}{\includegraphics[height=0.5in]{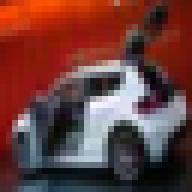}} & \adjustbox{center,valign=m,margin=1}{\includegraphics[height=0.5in]{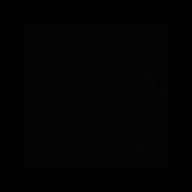}}& 
		0.992114 & 
		0.000466\\
		\hline
		bird&\adjustbox{center,valign=m,margin=1}{\includegraphics[height=0.5in]{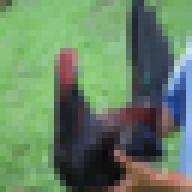}} & \adjustbox{center,valign=m,margin=1}{\includegraphics[height=0.5in]{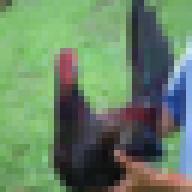}} & \adjustbox{center,valign=m,margin=1}{\includegraphics[height=0.5in]{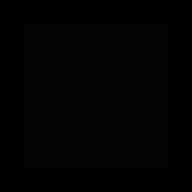}}& 
		0.959018 & 
	    0.001097\\
			\hline
		cat&\adjustbox{center,valign=m,margin=1}{\includegraphics[height=0.5in]{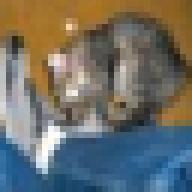}} & \adjustbox{center,valign=m,margin=1}{\includegraphics[height=0.5in]{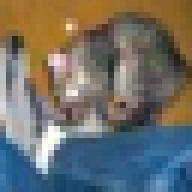}} & \adjustbox{center,valign=m,margin=1}{\includegraphics[height=0.5in]{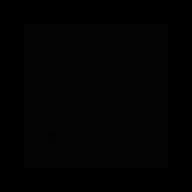}}& 
		0.983587 & 
		0.003116\\
				\hline
		deer&\adjustbox{center,valign=m,margin=1}{\includegraphics[height=0.5in]{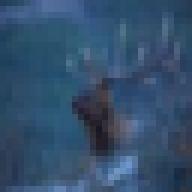}} & \adjustbox{center,valign=m,margin=1}{\includegraphics[height=0.5in]{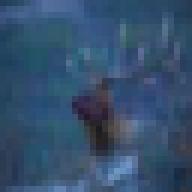}} & \adjustbox{center,valign=m,margin=1}{\includegraphics[height=0.5in]{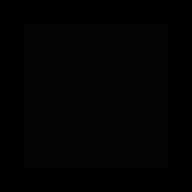}}& 
		0.947664 & 
		0.003062 \\
				\hline
		dog&\adjustbox{center,valign=m,margin=1}{\includegraphics[height=0.5in]{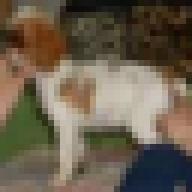}} & \adjustbox{center,valign=m,margin=1}{\includegraphics[height=0.5in]{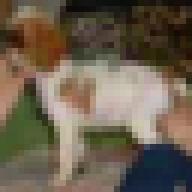}} & \adjustbox{center,valign=m,margin=1}{\includegraphics[height=0.5in]{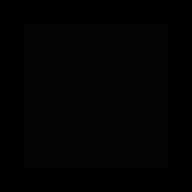}}& 
		0.985340 & 
		0.002234\\
				\hline
		frog&\adjustbox{center,valign=m,margin=1}{\includegraphics[height=0.5in]{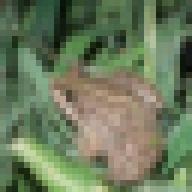}} & \adjustbox{center,valign=m,margin=1}{\includegraphics[height=0.5in]{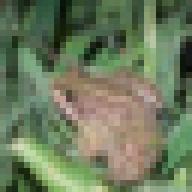}} & \adjustbox{center,valign=m,margin=1}{\includegraphics[height=0.5in]{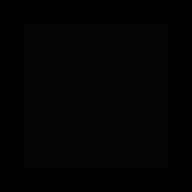}}& 
		0.983045 & 
		0.002367\\
				\hline
		horse&\adjustbox{center,valign=m,margin=1}{\includegraphics[height=0.5in]{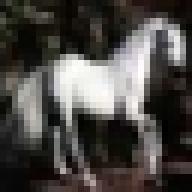}} & \adjustbox{center,valign=m,margin=1}{\includegraphics[height=0.5in]{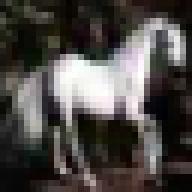}} & \adjustbox{center,valign=m,margin=1}{\includegraphics[height=0.5in]{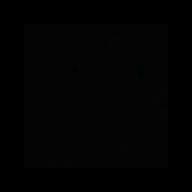}}& 
		0.995333 & 
		0.000964\\
				\hline
		ship&\adjustbox{center,valign=m,margin=1}{\includegraphics[height=0.5in]{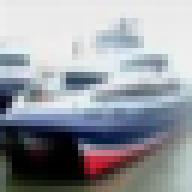}} & \adjustbox{center,valign=m,margin=1}{\includegraphics[height=0.5in]{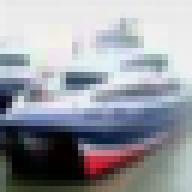}} & \adjustbox{center,valign=m,margin=1}{\includegraphics[height=0.5in]{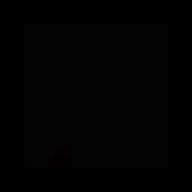}}& 
		0.982897 & 
		0.001897\\
				\hline
		truck&\adjustbox{center,valign=m,margin=1}{\includegraphics[height=0.5in]{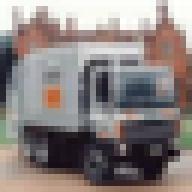}} & \adjustbox{center,valign=m,margin=1}{\includegraphics[height=0.5in]{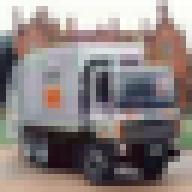}} & \adjustbox{center,valign=m,margin=1}{\includegraphics[height=0.5in]{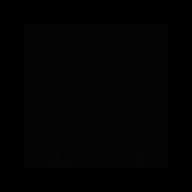}}& 
		0.990353 & 
		0.002192\\
		\hline
		\end{tabular}
\end{table}

The invisibility of the new trigger is also evaluated on the BTSR dataset in Table~\ref{tab:qual_comp}, where a clean image and its poisoned versions under different backdoor methods are provided. 
With the highest SSIM and the lowest LPIPS, the proposed trigger is the least visible among all the methods. Specifically, the LPIPS of the trigger is about 5, 20, and 1.5 times lower than those of the original BadNets, Trojaned NN, and HB, respectively (see the second, fourth, sixth, and eighth rows in Table~\ref{tab:qual_comp}).

We also investigate the relationship between the magnitude~$m$ and the invisibility of the proposed trigger by taking the CIFAR-10 dataset for example. As shown in Table~\ref{tab:relation_m_visibility}, even when the ASR of the trigger is as high as 96.49\% (even higher than the classification accuracy of benign images) under $m=6$, the new trigger is still invisible (and effective in terms of attack success).

    \begin{table}
    \caption{The imperceptibility of poisoned images on a $224\times 224$ BTSR image, where the images are resized for display. {\color{black}Trojaned NN can trojan ``1 neuron", ``2 neurons", or ``all neurons" of a selected layer in the DNN model. }}
    \label{tab:qual_comp} 
    \centering
    \begin{tabular}{m{4em}m{4em}m{4em}m{4em}m{3em}m{3em}}
    \hline 
    Method & Origin & Poisoned& Residual& SSIM& LPIPS \\ 
    \hline 
		\makecell{Ours \\ ($m=2$)} & 
		\adjustbox{center,valign=m,margin=1}{\includegraphics[height=0.5in]{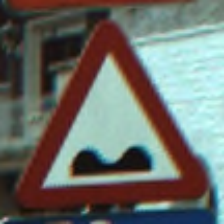}} &
		\adjustbox{center,valign=m,margin=1}{\includegraphics[height=0.5in]{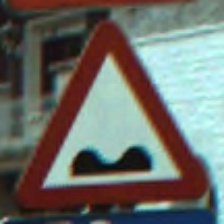}} & \adjustbox{center,valign=m,margin=1}{\includegraphics[height=0.5in]{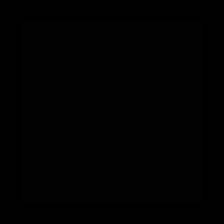}} & 
		\centering 0.993159 & 
		0.004201\\
		\hline
		\makecell{Ours  \\ ($m=3$)} & 
		\adjustbox{center,valign=m,margin=1}{\includegraphics[height=0.5in]{./img/clean_label_0_.png}} &
		\adjustbox{center,valign=m,margin=1}{\includegraphics[height=0.5in]{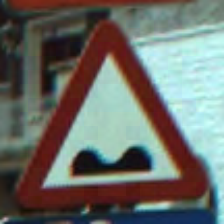}} & \adjustbox{center,valign=m,margin=1}{\includegraphics[height=0.5in]{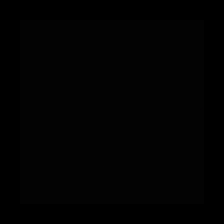}} & 
		0.986918 & 
		0.009344\\
		\hline
		\makecell{BadNets \\ (Standard,\\ $T=0$)} & 
		\adjustbox{center,valign=m,margin=1}{\includegraphics[height=0.5in]{./img/clean_label_0_.png}} &
		\adjustbox{center,valign=m,margin=1}{\includegraphics[height=0.5in]{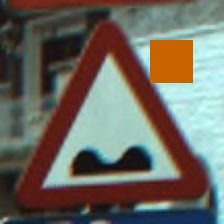}} & \adjustbox{center,valign=m,margin=1}{\includegraphics[height=0.5in]{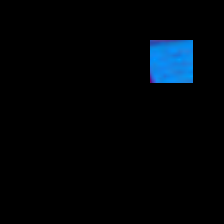}} & 
		0.965563 & 
		0.136998\\
		\hline
		\makecell{BadNets \\ ($T=0.7$)} & 
		\adjustbox{center,valign=m,margin=1}{\includegraphics[height=0.5in]{./img/clean_label_0_.png}} &
		\adjustbox{center,valign=m,margin=1}{\includegraphics[height=0.5in]{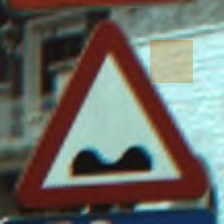}} & \adjustbox{center,valign=m,margin=1}{\includegraphics[height=0.5in]{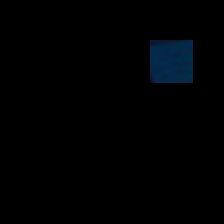}} & 
		0.988416 & 
		0.045952\\
		\hline
	    \makecell{Trojaned NN \\ (1 neuron,\\ $T=0$)}  & 
		\adjustbox{center,valign=m,margin=1}{\includegraphics[height=0.5in]{./img/clean_label_0_.png}} &
		\adjustbox{center,valign=m,margin=1}{\includegraphics[height=0.5in]{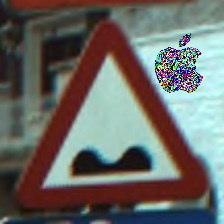}} & \adjustbox{center,valign=m,margin=1}{\includegraphics[height=0.5in]{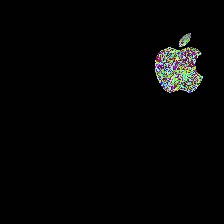}} & 
		0.946696 & 
		0.175602\\
		\hline
	    \makecell{Trojaned NN \\ (1 neuron,\\ $T=0.7$)}  & 
		\adjustbox{center,valign=m,margin=1}{\includegraphics[height=0.5in]{./img/clean_label_0_.png}} &
		\adjustbox{center,valign=m,margin=1}{\includegraphics[height=0.5in]{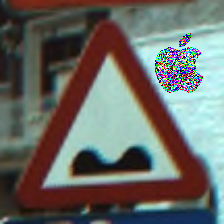}} & \adjustbox{center,valign=m,margin=1}{\includegraphics[height=0.5in]{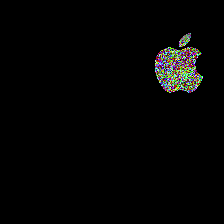}} & 
		0.946105 & 
		0.185182\\
		\hline
		\makecell{HB \\ (Standard,\\ $T=0$)} & 
		\adjustbox{center,valign=m,margin=1}{\includegraphics[height=0.5in]{./img/clean_label_0_.png}} &
		\adjustbox{center,valign=m,margin=1}{\includegraphics[height=0.5in]{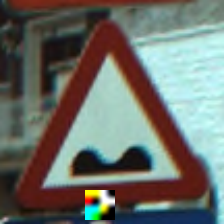}} & \adjustbox{center,valign=m,margin=1}{\includegraphics[height=0.5in]{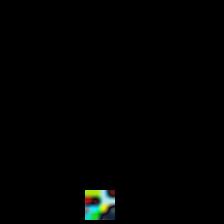}} & 
		0.976996 & 
		0.054661\\
		\hline
		\makecell{HB \\ ($T=0.7$)} & 
		\adjustbox{center,valign=m,margin=1}{\includegraphics[height=0.5in]{./img/clean_label_0_.png}} &
		\adjustbox{center,valign=m,margin=1}{\includegraphics[height=0.5in]{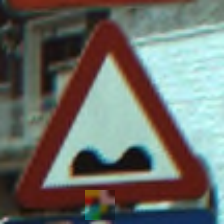}} & \adjustbox{center,valign=m,margin=1}{\includegraphics[height=0.5in]{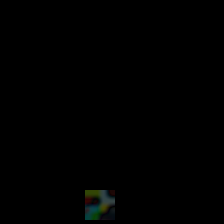}} & 
		0.985550 & 
		0.014404\\
		\hline
	\end{tabular}
	\end{table}

\begin{table}
\caption{{\color{black}The impact of perturbation magnitude on the visibility of the proposed trigger. The Functionality of the benign ResNet-18 model is 89.5\%. }}
\label{tab:relation_m_visibility} 
\centering
    \begin{tabular}{m{0.5em}m{4em}m{4em}m{3em}m{3em}m{3em}m{3em}}
    \hline 
     $m$ & Poisoned& Residual&ASR&Func.& SSIM& LPIPS \\ 
    \hline 
		1&\adjustbox{center,valign=m,margin=1}{\includegraphics[height=0.5in]{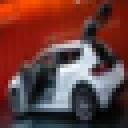}} & \adjustbox{center,valign=m,margin=1}{\includegraphics[height=0.5in]{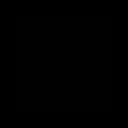}} & 92.48\%&89.01\% &
		0.991773 & 
	    0.000315\\
	    \midrule
		2&\adjustbox{center,valign=m,margin=1}{\includegraphics[height=0.5in]{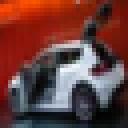}} & \adjustbox{center,valign=m,margin=1}{\includegraphics[height=0.5in]{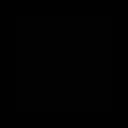}} & 91.34\%& 88.06\%&
		0.993684 & 
		0.000665\\
		\midrule
		3&\adjustbox{center,valign=m,margin=1}{\includegraphics[height=0.5in]{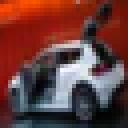}} & \adjustbox{center,valign=m,margin=1}{\includegraphics[height=0.5in]{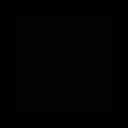}} &92.08\% & 89.94\%&
		0.991707 & 
		0.000752\\
		\midrule
		4&\adjustbox{center,valign=m,margin=1}{\includegraphics[height=0.5in]{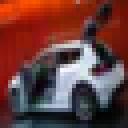}} & \adjustbox{center,valign=m,margin=1}{\includegraphics[height=0.5in]{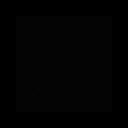}} & 95.30\%& 87.10\%&
		0.990609 & 
		0.001824\\
		\midrule
		5&\adjustbox{center,valign=m,margin=1}{\includegraphics[height=0.5in]{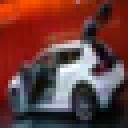}} & \adjustbox{center,valign=m,margin=1}{\includegraphics[height=0.5in]{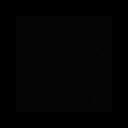}} &93.78\% &88.10\%&
		0.974187 & 
		0.001396\\
		\midrule
		6&\adjustbox{center,valign=m,margin=1}{\includegraphics[height=0.5in]{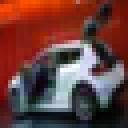}} & \adjustbox{center,valign=m,margin=1}{\includegraphics[height=0.5in]{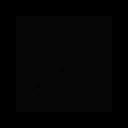}} &96.49\%&88.15\%&
		0.970179& 
		0.003235\\
		\hline
		\end{tabular}
 \end{table}

The superb imperceptibility of the new trigger is the result of the large dimension and subsequently the weak per-pixel perturbation of the trigger. Specifically, the perturbation of the new trigger is dispersed throughout large areas of a poisoned image. The perturbation to each individual pixel is weak, e.g., $m \leq 10$, substantially smaller than the maximum magnitude of 256 per RGB channel (i.e., less than 5\%). Moreover, the uniformly randomly produced elements inside the trigger can be viewed as noises to the images poisoned. With an adequate selection of $m$, the trigger can be imperceptible to human eyes while remaining highly effective in attack success.

\subsection{Attack Performance vs. Imperceptibility}
\begin{figure}
	    \centering
	    \includegraphics[height=2in]{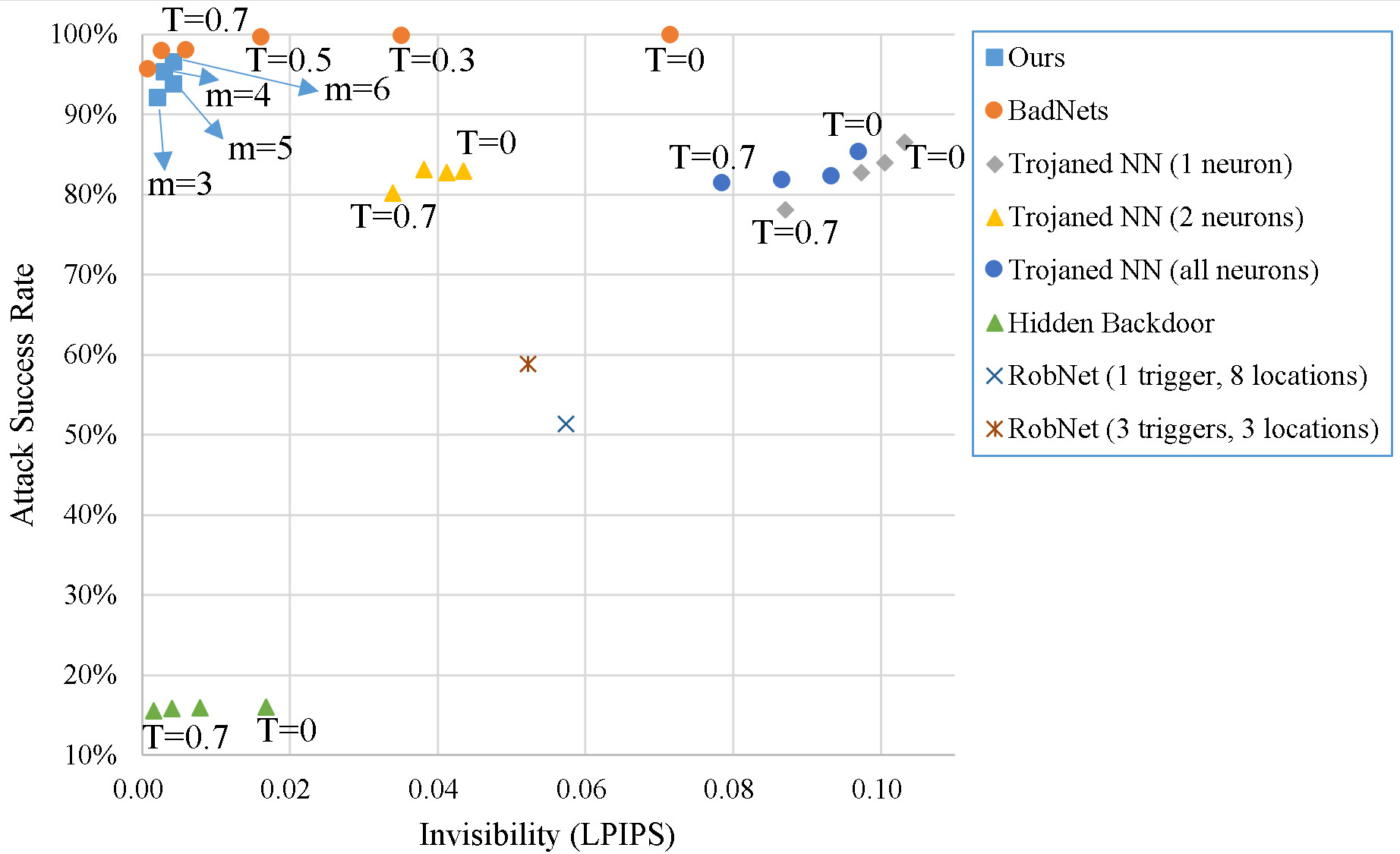}
	    \caption{{\color{black}The trade-off of ASR and invisibility of triggers on the CIFAR-10 dataset. The invisibility is measured by LPIPS. A lower LPIPS indicates better invisibility.}}
	    \label{fig:asr_lpips_CIFAR-10}
	\end{figure}

\begin{figure}
	    \centering
	    \includegraphics[height=2in]{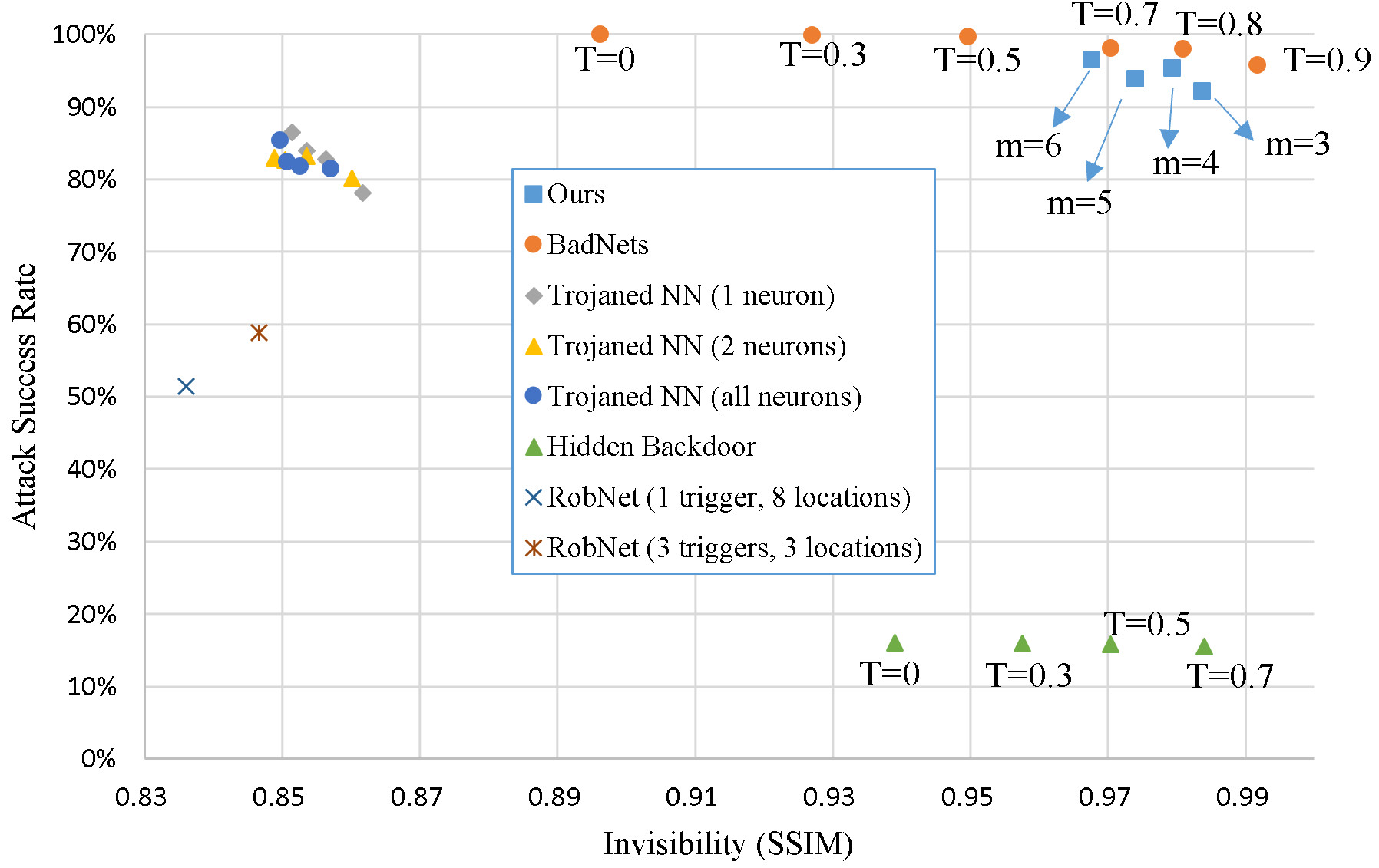}
	    \caption{{\color{black}The trade-off of ASR and invisibility of triggers on the CIFAR-10 dataset. The invisibility is measured by SSIM. A higher SSIM indicates better invisibility.}}
	    \label{fig:asr_ssim_CIFAR-10}
	\end{figure}

\begin{figure}
	    \centering
	    \includegraphics[height=2in]{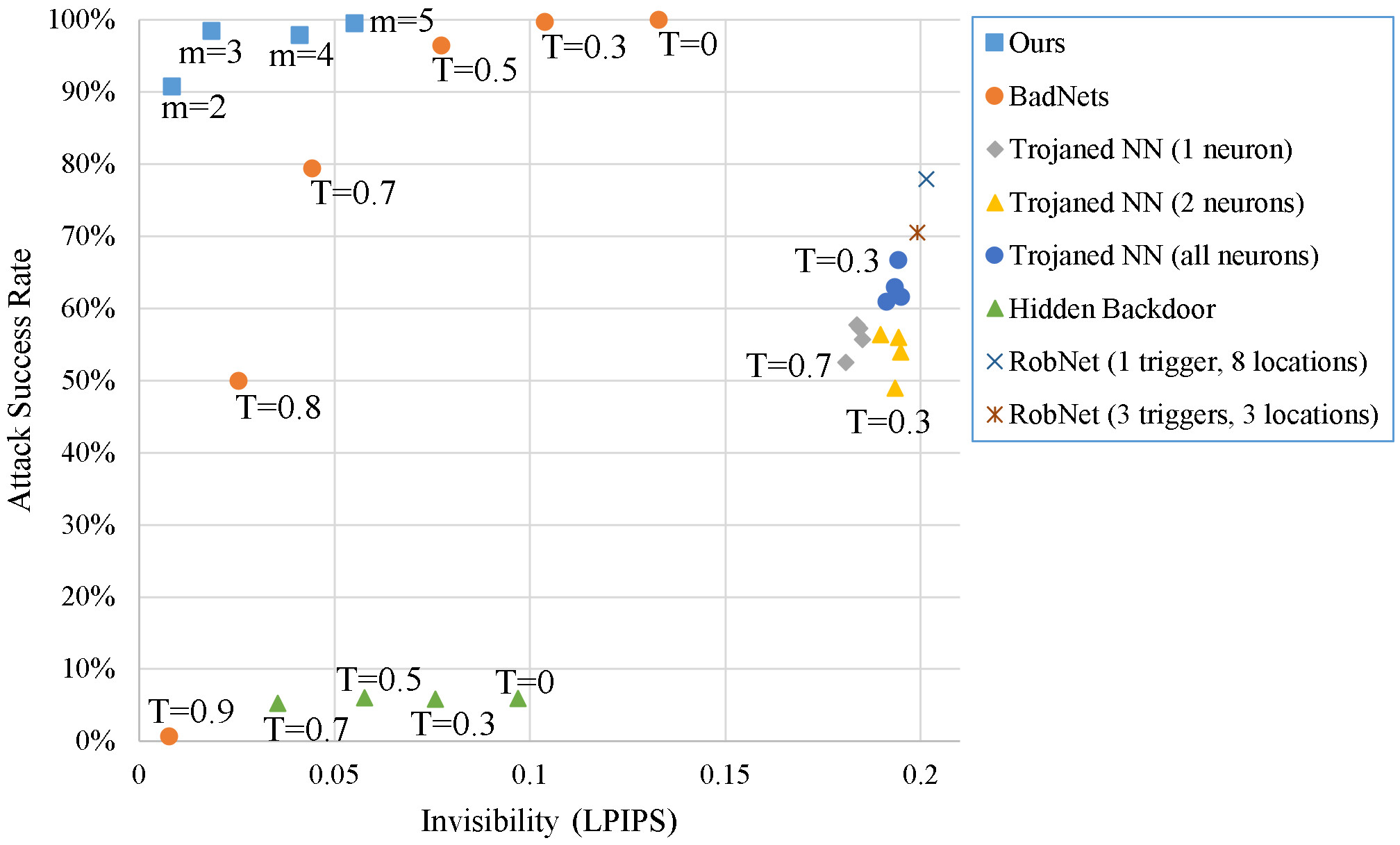}
	    \caption{{\color{black}The ASR vs. invisibility (measured by LPIPS) of the considered triggers on the BTSR dataset. }}
	    \label{fig:asr_lpips_btsr}
	\end{figure}

\begin{figure}
	    \centering
	    \includegraphics[height=2in]{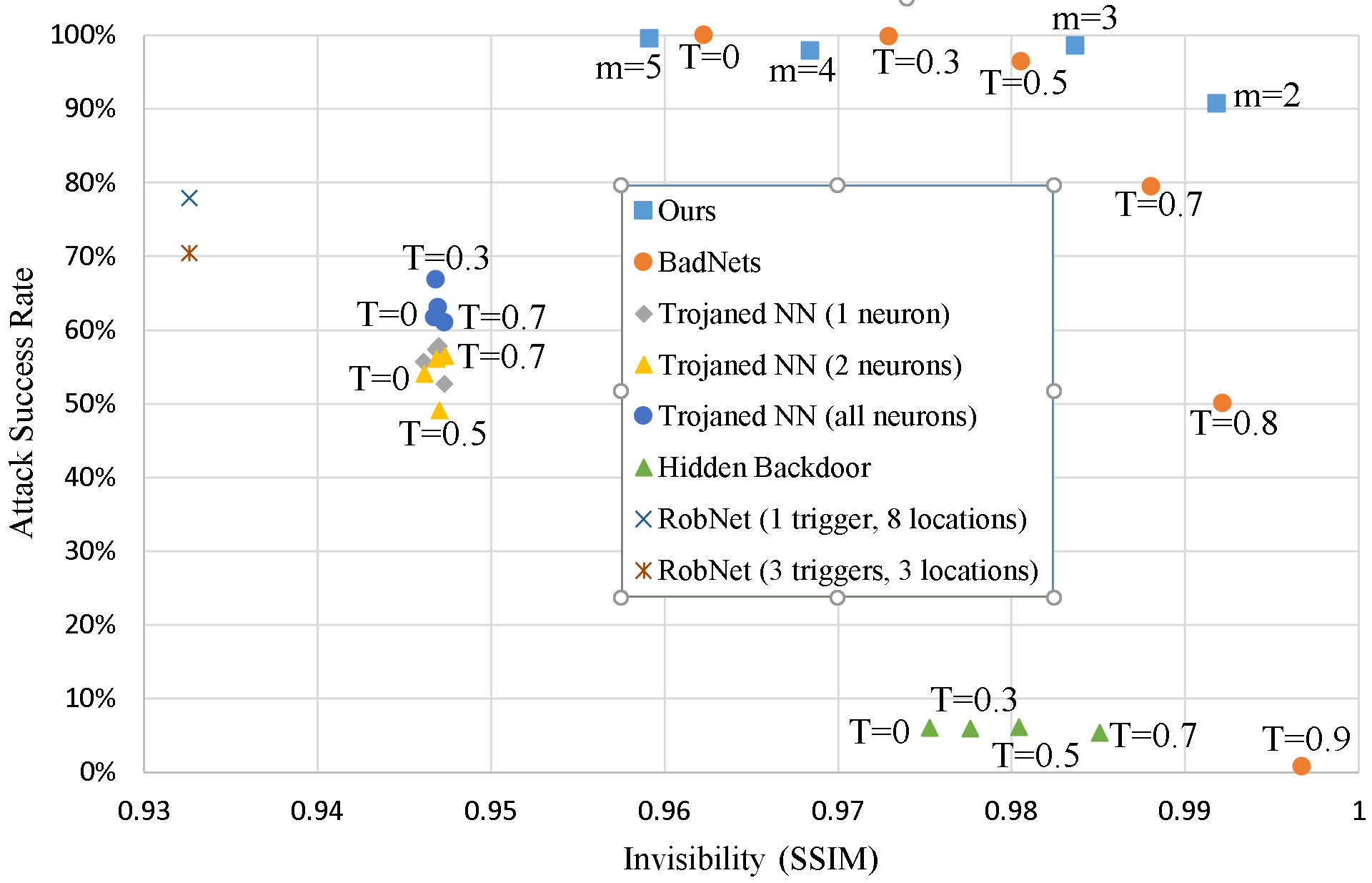}
	    \caption{{\color{black}The trade-off of ASR and invisibility (measured by SSIM) of the considered triggers on the BTSR dataset.}}
	    \label{fig:asr_ssim_bstr}
	\end{figure}

	Figs.~\ref{fig:asr_lpips_CIFAR-10}--\ref{fig:asr_ssim_bstr} show the trade-off between the ASR and invisibility of backdoor triggers assessed on the CIFAR-10 and BTSR dataset. The invisibility is measured by LPIPS and SSIM. For a comprehensive comparison, the transparency of a trigger is adjusted by configuring a transparency parameter $0 \leq T < 1$ in BadNets and HB, as in Trojaned NN~\cite{DBLP:conf/ndss/LiuMALZW018}. The trigger is opaque if $T=0$. It is more transparent if $T$ is larger.
	
	Figs.~\ref{fig:asr_lpips_CIFAR-10} and~\ref{fig:asr_ssim_CIFAR-10} show that the new trigger and BadNets perform significantly better than the other considered methods on the CIFAR-10 dataset. The results of the new trigger and BadNets are localized in the upper left corner of Fig.~\ref{fig:asr_lpips_CIFAR-10} and the upper right corner of Fig.~\ref{fig:asr_ssim_CIFAR-10}, indicating the new trigger and BadNets can achieve both high ASRs and imperceptibility (high in SSIM and low in LPIPS). 
	It is worth pointing out that the original design of BadNets only uses opaque triggers (i.e., $T=0$). While the extended BadNets with transparent triggers can marginally outperform the new trigger, the original BadNets performs poorly in imperceptibility.  
	Moreover, the new trigger generally has a smaller Functionality Loss than BadNets on the CIFAR-10 dataset, as revealed in Table~\ref{tbl:func_loss_compare}. 

	The superiority of the new trigger to the other methods, including BadNets, is revealed on the BTSR dataset in terms of both ASR and invisibility, as shown in Figs.~\ref{fig:asr_lpips_btsr} and~\ref{fig:asr_ssim_bstr}. The images in the BTSR dataset have a larger dimension (i.e., $224\times 224$ pixels) than those in the CIFAR-10 dataset (i.e., $32 \times 32$ pixels). The larger dimension of the images allows for a larger size and smaller perturbation magnitude of the new trigger, benefiting both attack success and imperceptibility.
	
\begin{table}
 \centering
 \caption{{\color{black}The comparison of Functionality Loss between the proposed trigger and BadNets on the CIFAR-10 dataset. The Functionality of the benign ResNet-18 model is 89.5\%.}}
 \renewcommand{\arraystretch}{1.2}
 \renewcommand\tabcolsep{2.5pt}
 \begin{tabular}{c|c|c|c|c|c|c}
  \hline
  Method &Parameter& Func.(\%) &Loss (\%)& ASR(\%) & SSIM & LPIPS\\
  \hline
  & $T=0$  &  82.79& 6.71&  99.91&0.896295 & 0.071578\\
  & $T=0.3$ &87.00 &  2.50& 99.83&0.927061  &0.035184 \\
  & $T=0.5$ &87.01 &  2.49& 99.58&0.949753 &0.016209 \\
  BadNets & $T=0.7$ &82.42 &  7.08& 98.04&0.970536 &0.005937 \\
  & $T=0.9$ &86.65 &  2.85& 95.65&0.991857 &0.000805\\
  \hline
  & $m=3$ &88.94 & 0.56& 92.08&0.983782 &0.002213\\
  Ours & $m=4$ &87.10 & 2.40& 95.30&0.979386 &0.003094\\
  & $m=5$ &88.10 & 1.40& 93.78&0.974112 &0.004347\\
  & $m=6$ &88.15 & 1.35& 96.49&0.967717 &0.004371\\
  \hline
 \end{tabular}
 \label{tbl:func_loss_compare}
\end{table}
	
We also compare the new backdoor trigger with RobNet~\cite{9450029}, an extension to Trojaned NN~\cite{DBLP:conf/ndss/LiuMALZW018} by supporting multiple locations for a trigger (up to eight different locations on an image) or multiple different triggers (one per location) for a backdoored model. 
	We adopt the setting of~\cite{9450029} on the trigger number and locations of RobNet: In the case of multiple locations for a trigger, we generate two poisoned versions of every clean image, each placing the trigger at a different and randomly selected location from the eight candidate locations. In the case of multiple triggers, we generate three poisoned versions of a clean image, each poisoned with a different trigger at a randomly selected location. 
	The poisoning rate (i.e., the ratio of poisoned images to all input images) of RobNet is set to be no lower than the poisoning rate of the new trigger. We set the poisoning rate of the new trigger to 5\%, which corresponds to 228 and 2500 poisoned images on the BTSR dataset and the CIFAR-10 dataset, respectively.  
	
	As shown in Table~\ref{tab:compare_with_robnet}, the new trigger achieves higher ASRs and significantly lower Functionality Loss than RobNet in all considered scenarios. 
	On the BTSR dataset, the trigger achieves the ASR of over 90\% with at most a 3.16\% loss of Functionality. In contrast, RobNet undergoes around 50\% loss of Functionality, and its ASR is just about 16\%.
	On the CIFAR-10 data, the highest ASR achieved by RobNet is 58.86\% at a 8.79\% Functionality Loss. By setting $m$ to be as small as $m=3$, the new trigger can achieve higher ASRs with much smaller Functionality Loss than RobNet. Moreover, the new trigger is less visible (with higher SSIM and lower LPIPS) than RobNet.
	
	\begin{table*}
    \caption{{\color{black}Comparison between the proposed method and RobNet. ResNet-18 is used as the neural network model.}}
    \label{tab:compare_with_robnet}
    \centering
    \begin{tabular}{c|c|c|c|c|c|c}
    		    \hline
    			{Dataset}&{Method}&{\#Poison}&{Attack Success Rate}  & {Functionality Loss} &{SSIM}& {LPIPS}\\
    			\hline   
    			&RobNet (8 locations)&$2\times114$&	0.28\% &46.40\%&0.932617&0.201440 \\
    			{BTSR}
    			&RobNet (3 triggers)&$3\times91$&	15.82\% &46.33\%&0.932605&0.199078 \\
    			&ours ($m=2$)&$228$&	90.74\% &3.12\%&0.991851&0.008482 \\
    			&ours ($m=3$)&$228$&	98.58\% &2.30\%&0.983669&0.018595 \\
    			&ours ($m=4$)&$228$&	97.94\% &2.14\%&0.968385&0.041093 \\
    			&ours ($m=5$)&$228$&	99.56\% &3.16\%&0.959122&0.055185 \\
    			\hline
    			&RobNet (8 locations)&$2\times1250$&51.27\% &6.80\%&0.835987&0.057399 \\
    		    {CIFAR-10}	
    		    &RobNet (3 triggers)&$3\times1000$&58.86\% &8.79\%&0.846650&0.052198 \\
    			&ours ($m=3$)&$2500$&	92.08\% &0.56\%&0.983782 
& 0.002213\\
    			&ours ($m=4$)&$2500$&	95.30\% &2.40\%&0.979386& 0.003094\\
    			&ours ($m=5$)&$2500$&	93.78\% &1.40\%&0.974112&0.004347 \\
    			&ours ($m=6$)&$2500$&	96.49\% &1.35\%&0.967717& 0.004371\\
    			\hline
    \end{tabular}
    \end{table*}

We also compare the proposed trigger with the input-agnostic adversarial trigger generation approaches, i.e., AdvGAN~\cite{DBLP:conf/ijcai/XiaoLZHLS18} and UAT~\cite{DBLP:conf/aaai/ShafahiN0DDG20}. As shown in Table~\ref{tbl:compare_with_adv}, the new trigger achieves significantly higher ASRs than AdvGAN and UAT on all the considered datasets. Take the CIFAR-10 dataset for an example. The new trigger outperforms AdvGAN and UAT by 17.26\% and 30.53\%, respectively. Considering the LPIPS metric, the new trigger is the most invisible on the MNIST dataset, less visible than AdvGAN on the CIFAR-10 dataset, and less visible than UAT on the BTSR dataset. Considering the SSIM metric, the new trigger is the most invisible on both the CIFAR-10 and BSTR datasets, and less visible than UAT on the MNIST dataset.

\begin{table}
 \centering
 \caption{{\color{black}Comparison of the proposed trigger with the existing input-agnostic, imperceptible attacks, AdvGAN~\cite{DBLP:conf/ijcai/XiaoLZHLS18} and UAT~\cite{DBLP:conf/aaai/ShafahiN0DDG20}. For AdvGAN, the number of epochs is 60 based on experimental tests to maximize the ASR (as no value was recommended in~\cite{DBLP:conf/ijcai/XiaoLZHLS18}). For UAT, the number of epochs is 10, as suggested in~\cite{DBLP:conf/aaai/ShafahiN0DDG20}.}}
 \renewcommand{\arraystretch}{1.2}
 \renewcommand\tabcolsep{2.5pt}
 {\color{black}
 \begin{tabular}{c|l|c|c|c}
  \hline
  &Method& ASR (\%)&SSIM& LPIPS\\
  \hline
  MLP&Ours ($m=10$)&96.80 &0.863732&0.000553\\
  on&AdvGAN ($\epsilon =0.3$)& 58.20&0.922047&0.180812\\
  MNIST&UAT ($\epsilon = 16/255$)&14.60&0.848677&0.097013\\
  \hline
  ResNet-18&Ours ($m=5$)&93.78 &0.974112&0.004347\\
  on&AdvGAN ($\epsilon =8/255$)&76.52&0.925907&0.009442\\
  CIFAR-10&UAT ($\epsilon = 10/255$)&63.25&0.906396&0.003110\\
  \hline
  ResNet-18&Ours ($m=5$)&99.56 &0.979386&0.003094\\
  on&AdvGAN ($\epsilon =10/255$)&24.33&0.967096&0.000045\\
  BTSR&UAT ($\epsilon = 10/255$)&24.22 &0.935355&0.104915\\
  \hline
 \end{tabular}
 }
 \label{tbl:compare_with_adv}
\end{table}

\subsection{Resistance to Existing Defense Methods}
\label{sec: resist}
{\color{black}
    One possible defense method is that a defender could decide to enumerate all possible 3D patterns (or realizations) of the trigger. 
	A poisoned image could be potentially detected by correlating the image with every possible realization of the trigger. Nevertheless, the use of the CSPRNG ensures consistently low correlations between any two different random sequences. The trigger would only be revealed if the same trigger is picked up for correlation.
	A backdoored DNN model could be potentially detected by perturbing (labeled) benign images with each possible trigger realization and inputting the perturbed images into the model to gauge the misclassification rate. An image perturbed by the real trigger would be classified to a different class from its correct class, and the misclassification rate increases. 
	The complexity of enumerating all possible trigger realizations grows exponentially with the elements in the trigger and is computationally prohibitive in practice.
}

Many defense algorithms have been proposed to counteract backdoored DNNs~\cite{DBLP:conf/acsac/GaoXW0RN19, DBLP:conf/acsac/DoanAR20,DBLP:conf/nips/Tran0M18,DBLP:conf/aaai/ChenCBLELMS19,DBLP:conf/ndss/Xu0Q18,DBLP:conf/sp/WangYSLVZZ19,DBLP:conf/ccs/LiuLTMAZ19,DBLP:conf/raid/0017DG18}, which can detect or mitigate backdoor triggers.
The new trigger is tested against eight recently published defense methods, namely, STRIP~\cite{DBLP:conf/acsac/GaoXW0RN19}, SSD~\cite{DBLP:conf/nips/Tran0M18}, AC~\cite{DBLP:conf/aaai/ChenCBLELMS19}, Februus~\cite{DBLP:conf/acsac/DoanAR20}, Neural Cleanse~\cite{DBLP:conf/sp/WangYSLVZZ19}, Spatial Smoothing~\cite{DBLP:conf/ndss/Xu0Q18}, ABS~\cite{DBLP:conf/ccs/LiuLTMAZ19}, and Neuron Pruning~\cite{DBLP:conf/raid/0017DG18}. 
We show that the trigger can escape the detection and scrutiny of the methods, and pose significant threats to image classification neural networks. 
The details of the eight state-of-the-art defense methods are provided in Section~\ref{sec: defenses}. 

\subsubsection{STRIP}
In this strategy, a backdoor trigger is detected by comparing the outputs of the backdoored DNN after being fed with clean and poisoned samples.
For the CIFAR-10 and MNIST datasets, we randomly select 2,000 images from a pool of 10,000 testing images and organize them into a clean group. The images in the clean group are then duplicated and poisoned with the new trigger to form a poisoned group. We superpose clean images from other classes than the selected 2,000 images, to both the clean and poisoned groups. For the BTSR dataset, we randomly select 1,000 images since there are a limited number of testing images.

\begin{figure}
	\centering
	\begin{subfigure}{0.35\textwidth}
		\includegraphics[width=2.5in]{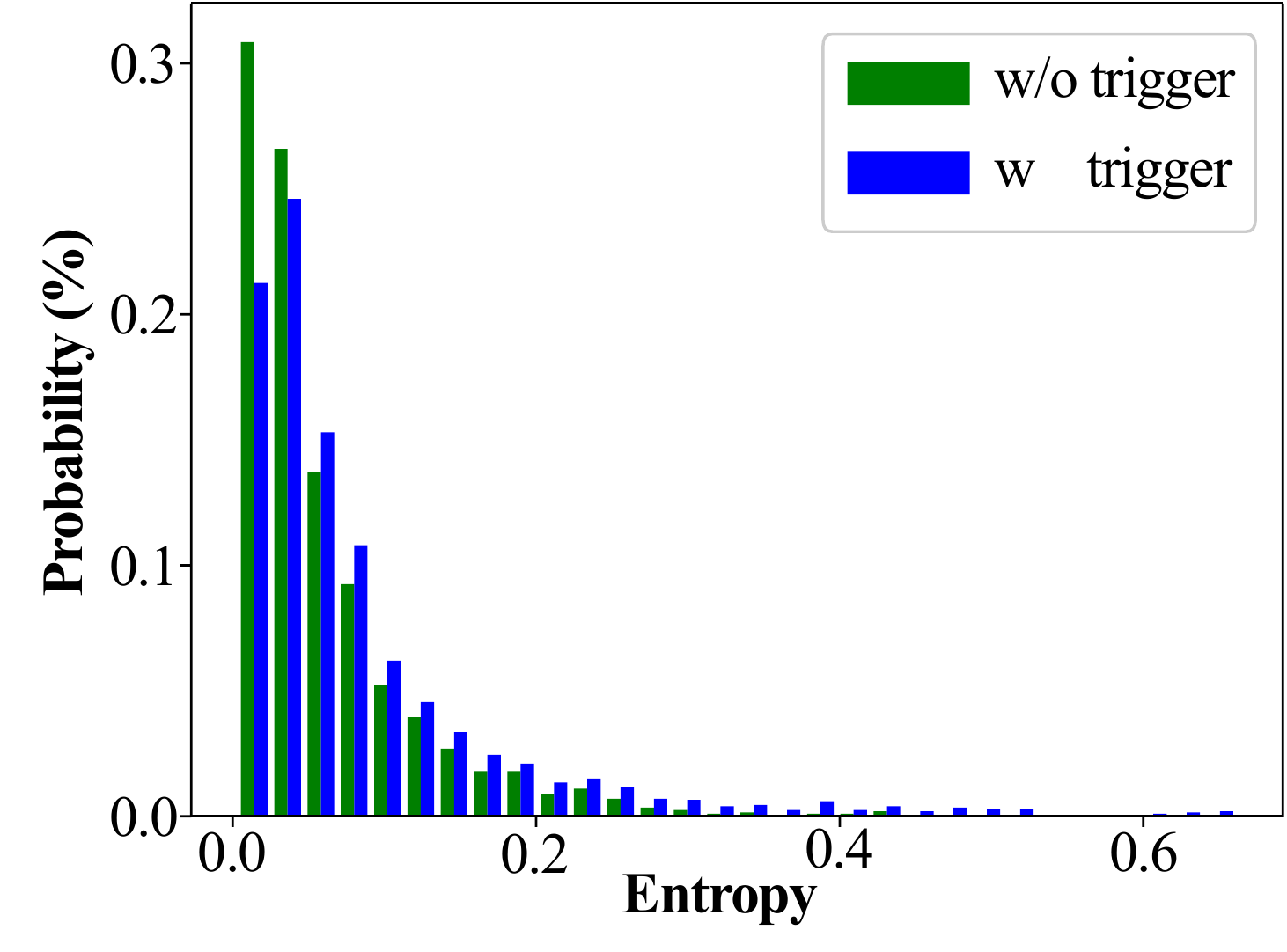}
		\caption{ResNet-18 on CIFAR-10}
	\end{subfigure}
	~
	\begin{subfigure}{0.35\textwidth}
		\includegraphics[width=2.5in]{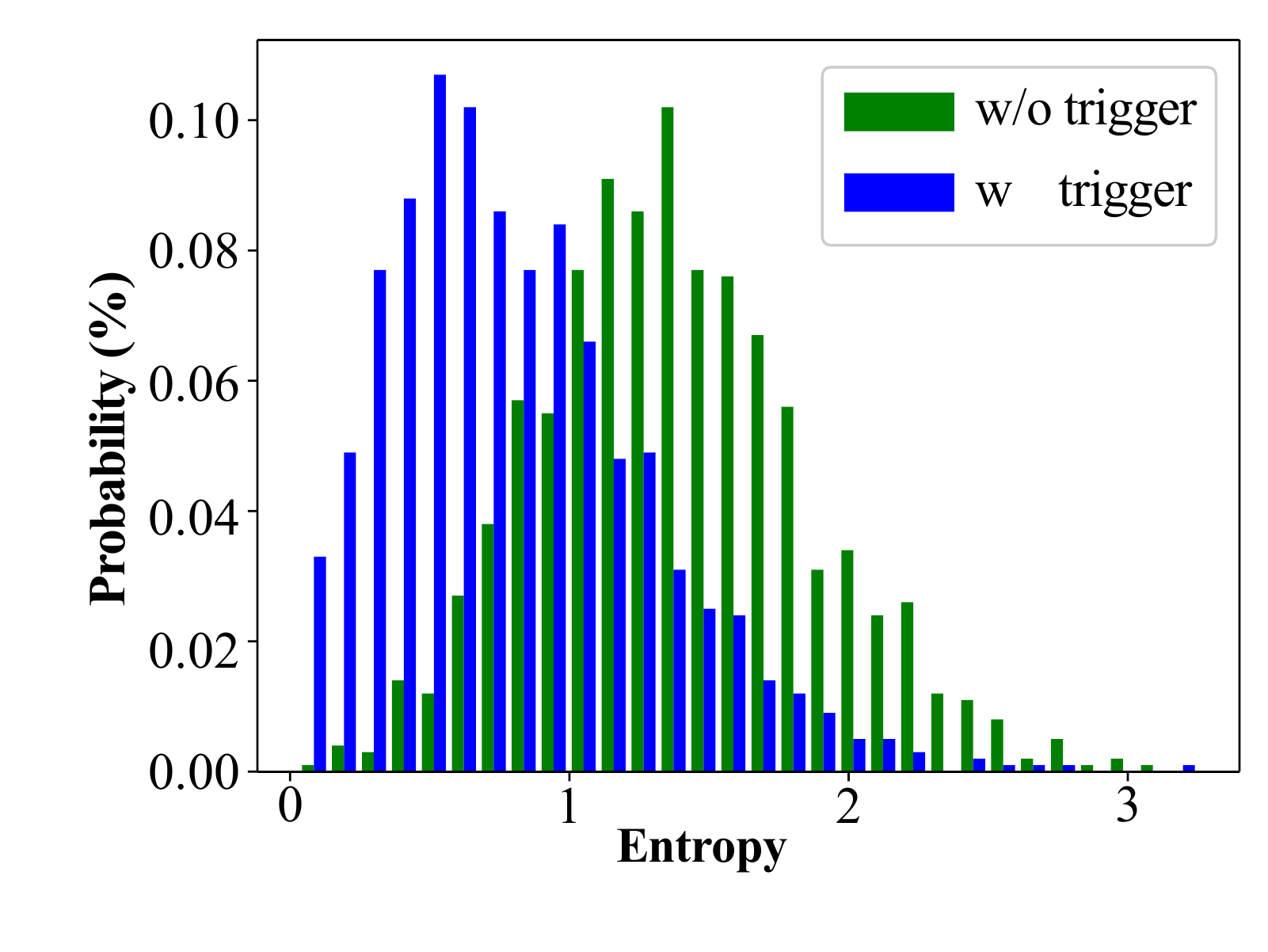}
		\caption{ResNet-18 on BTSR}
	\end{subfigure}
	~
	\centering
	\begin{subfigure}{0.35\textwidth}
		\includegraphics[width=2.5in]{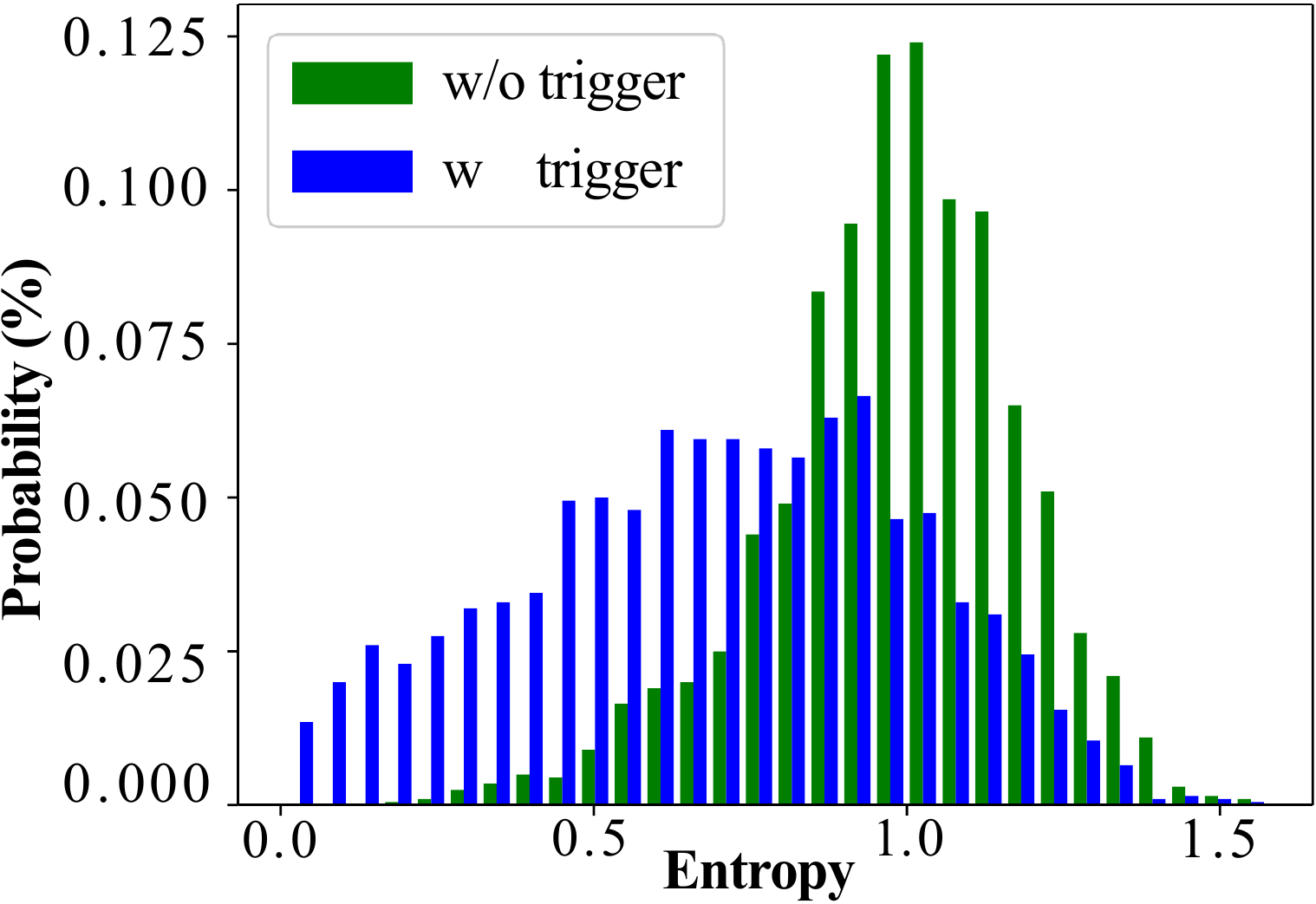}
		\caption{MLP on MNIST}
	\end{subfigure}
	\caption{The STRIP results on MNIST, CIFAR-10 and BTSR. The distributions of the entropy of samples with and without the new trigger overlap substantially, making setting a detection threshold impossible.}
	\label{fig: result-strip}
\end{figure}

Fig.~\ref{fig: result-strip} plots the histogram of entropy. 
The entropy of samples with and without the new backdoor trigger has nearly the same distribution in Fig.~\ref{fig: result-strip}(a). The entropy of both clean and poisoned CIFAR-10 samples is primarily lower than 0.2, and more than half of them are close to zero. 
In Figs.~\ref{fig: result-strip}(b) and~\ref{fig: result-strip}(c), while the distributions of the entropy of the samples with and without the backdoor trigger are different,  
the entropy of most samples in the BTSR and MNIST datasets ranges from 0.5 to 1.5, and from 0.5 to 1.3, respectively. It is difficult to derive a threshold to separate the clean and poisoned samples based on the entropy. 

\subsubsection{Spectral Signature Defense and Activation Clustering}	
SSD~\cite{DBLP:conf/nips/Tran0M18} and AC~\cite{DBLP:conf/aaai/ChenCBLELMS19} examine the difference of internal representations (i.e., activations) between clean and poisoned images in the hidden layers (e.g., penultimate layer) of a DNN. 

\begin{table*}[!ht]
	\caption{The detection result of SSD~\cite{DBLP:conf/nips/Tran0M18} and AC~\cite{DBLP:conf/aaai/ChenCBLELMS19}.}
	\label{tab: ssd}
	\centering
\begin{tabular}{l|l|r|r|r|r|r|r|r|r|r}
		\hline
		\multicolumn{2}{c}{Method}\vline&\multicolumn{3}{c}{{MLP on MNIST}}\vline&\multicolumn{3}{c}{{ResNet18 on CIFAR10}}\vline&\multicolumn{3}{c}{{RetNet18 on BTSR}}\\
		\multicolumn{2}{c}{}\vline&{TPR} &{TNR}&{bACC}&{TPR} &{TNR}&{bACC}&{TPR} &{TNR}&{bACC}\\
		\hline   
		\multicolumn{2}{c}{SSD}\vline& 0.920& 0.081&0.501& 0.926&  0.080	&0.503&0.996&0&0.498\\
		\cline{2-11}
		\hline
		&{PCA + Smaller}& 0.494&0.572& 0.533& 0.027&  0.706&0.367&0.407&0.751&0.579\\
		&{PCA + Distance}& 0& 1&0.500& 0.027&  0.104&	0.066&0.422&0.989&0.706\\
		&{PCA + RelativeSize}&0 & 1&0.500& 0.027&  0.827&0.427&0&0.858&0.429 		\\
		&{PCA + Silhouette}& 0& 1&0.500& 1&  0.2&0.600&0&0.365&0.183 			\\
		&{FastICA + Smaller}& 0.494&0.572 &0.533& 0.024&  0.787&0.406&0.407&0.715&0.561 			\\
		{AC}&{FastICA + Distance}& 0&1 &0.500& 0.014&  0.830&	0.422&0.411&0.980&0.696 		\\
		&{FastICA + RelativeSize}&0 & 1&0.500& 0.014&  0.820&0.417&0&0.835&0.418 			\\
		&{FastICA + Silhouette}&0 &1 &0.500& 1&  0&	0.500&0&0.172&0.086 		\\
		&{TSNE + Smaller}& 0.494&0.573 &0.534& 0.028&  0.708&0.368&0.398&0.741&0.570 			\\
		&{TSNE + Distance}&0 &1 &0.500& 0.028&  0.960&0.494&0.411&0.985&0.698 		\\
		&{TSNE + RelativeSize}& 0&1 &0.500& 0.028&  0.829&0.429&0&0.842&0.421 			\\
		&{TSNE + Silhouette}&0 & 1&0.500& 1&  0.301&0.651&0&0.210&0.105 			\\
		\cline{2-11}
		\hline
	\end{tabular}
\end{table*}

Table~\ref{tab: ssd} shows that SSD has a significantly lower {\color{black}TNR} than {\color{black}TPR} on all the three datasets, with the TNRs lower than 0.1 for all the considered values of $\epsilon_m$. This is because SSD first determines the trigger detection threshold $\xi$ by~\eqref{eqn: ssd_poison}, and then confirms clean samples if $\langle X - \mu_\mathcal{F}, v \rangle > \xi$ in~\eqref{eqn: ssd_clean}. As samples poisoned by the new trigger do not satisfy the $\epsilon$-spectrally separable condition described in Section~\ref{sec: defenses}, most clean samples yield $\langle X - \mu_\mathcal{F}, v \rangle < \xi$. As a result, SSD classifies nearly all samples as poisoned images and obtains a bACC of around 0.5. SSD cannot detect the trigger.

Table~\ref{tab: ssd} also shows that the bACC of AC is less than 0.7 on the different datasets, while most of its TPRs are much lower than the TNRs. The TPR and TNR of the AC method vary significantly across different parameter configurations and datasets. On the low-dimensional grey-scale MNIST images, using the Smaller method (see Section~\ref{sec: defenses}) with AC allows for a TPR of 0.494 and a TNR of 0.572. The rest of the methods cause AC to misclassify all samples to be clean images and obtain a TPR of zero. On the high-dimensional color BTSR images, the Distance method (see Section~\ref{sec: defenses}) allows AC to achieve a high TNR between 0.980 and 0.989 along with a TPR ranging from 0.411 to 0.422, resulting in a small bACC value ranging from 0.696 to 0.706.

Techniques that reduce the dimensions of images, such as PCA, FastICA, and $t$-SNE, cause little difference. AC fails to detect the new trigger, as $K$-Means is unsuitable for high-dimensional data while the dimension reduction of the activations can compromise the images. Other clustering methods, such as DBSCAN, Gaussian Mixture Models, and Affinity Propagation, perform worse in clustering dimension-reduced activations than $K$-Means~\cite{DBLP:conf/aaai/ChenCBLELMS19}. Because the new trigger has a smaller perturbation magnitude than a visible trigger (e.g., the yellow square in~\cite{DBLP:journals/corr/abs-1708-06733}), it gets obfuscated during the dimension reduction. As a result, it is difficult to distinguish between poisoned and clean samples using the reduced activations in AC.

\subsubsection{Februus}
Under the assumption that the pixels corresponding to the activation hot spots in a DNN's penultimate layer are potentially a backdoor trigger, Februus removes the trigger by replacing the pixels with an image patch recovered by a pre-trained GAN model~\cite{DBLP:conf/acsac/DoanAR20}. 
We reuse the model architecture and pre-trained GAN model provided in~\cite{DBLP:conf/acsac/DoanAR20}. On the CIFAR-10 dataset, a DNN with six convolution layers and two dense (i.e., fully-connected) layers is trained with clean images and images poisoned with the new trigger. 

\begin{table}
		\caption{Comparison of Functionality and ASR before and after applying Februus.}
		\label{tab: februus}
		\centering
		\begin{tabular}{c|c|r|r|r|r}
			\hline
			{Model} &{Dataset} &\multicolumn{2}{c}{{Before}}\vline&\multicolumn{2}{c}{{After}}\\
			&&{Func.} &{ASR} &{Func.} &{ASR}\\
			\hline   
			{6 Conv + 2 Dense}&{CIFAR-10}& 87.8\% &92.2\% &42.5\% & 18.8\%\\
			{ResNet-18}&{BTSR}& 91.8\% &98.7\% &91.4\% & 98.6\% \\
			\hline
		\end{tabular}
\end{table}

Table~\ref{tab: februus} shows that Functionality and ASR are 87.8\% and 92.2\%, respectively. Although the ASR is reduced to 18.8\% by applying Februus, the Functionality of the network is also reduced significantly to 42.5\%, rendering the model useless. On the BTSR dataset, the ResNet-18 achieves the Functionality of 91.8\% and the ASR of 98.7\% before the trigger removal. However, the Functionality drops to 91.4\% after the trigger removal, while the ASR remains barely changed. 
On the CIFAR-10 dataset, the significant reduction in the Functionality is due to the fact that poisoned samples with the new trigger produce similar hot spots to clean samples and Februus using CAM cannot tell their difference. 

As shown in Fig.~\ref{fig: cam-CIFAR-10}, the hot spot (shaded in red) in a poisoned image overlaps with the object to be classified (e.g., the majority of the ship body). Furthermore, because the pixels outside the hot spot are mainly the background and provide little information about the original image, the GAN model cannot recover the original clean image patch at the hot spot. The barely changed ASR on the BTSR dataset is due to the fact that the magnitude used is too weak for CAM to produce a meaningful hot spot. Few pixels are changed by the pre-trained GAN used in Februus. As a result, the trigger cannot be removed without compromising the classification capability (i.e., the Functionality) of a DNN. 

\begin{Figure}
	\centering
	\includegraphicsmaybe[width=7cm]{img/cam-CIFAR10.pdf}
	\captionof{figure}{An illustration of the CAM of samples in the CIFAR-10 dataset, where the red region is the hot spot whose pixels contribute the most to the image classification.}
	\label{fig: cam-CIFAR-10}
\end{Figure}

\subsubsection{Neuron Pruning}
Table~\ref{tab:prune_result} shows the resistance of the new backdoor trigger to Neuron Pruning, where the threshold of terminating Neuron Pruning is over 4\% reduction in Functionality Loss (i.e., the  loss of classification accuracy on clean inputs), as considered in~\cite{DBLP:conf/raid/0017DG18}.
It is shown that on the BTSR dataset, Neuron Pruning can only reduce the ASR of the proposed trigger by only 3.76\% (from 99.13\% to 95.37\%), which is even smaller than the 4.43\% decrease in the classification accuracy. On the CIFAR-10 dataset, Neuron Pruning even increases the ASR by 2.55\%. 
On the low-dimensional gray-scale MNIST dataset, the ASR drops by 22.5\% for a shallow MLP model. However, the residual ASR is still as high as 74.26\%, posing a considerable threat to safety- or security-critical applications.

\begin{table}
    \centering
    \caption{{\color{black}The new trigger under Neuron Pruning.}}
    \begin{tabular}{c|c|c|c|c|c}
        \hline
         Model& Dataset &ASR&ASR& ASR& Func.\\
             &&before&after&Change&Loss \\
        \hline
         MLP &MNIST  & 96.76\%& 74.26\%&  -22.5\% & 4.13\% \\
       	 \hline 	         
         ResNet-18&BTSR  &99.13\% & 95.37\%& -3.76\% & 4.43\%\\
        \hline
        ResNet-18 &CIFAR-10 & 95.28\%& 97.83\%& +2.55\% &4.09\%\\
        \hline
    \end{tabular}
    \label{tab:prune_result}
\end{table}

\subsubsection{Neural Cleanse} 
Given a potentially backdoored neural network model, Neural Cleanse obtains a set of triggers, each of which can cause misclassification to the corresponding target label and has the minimal footprint in terms of the number of occupied pixels and color intensities~\cite{DBLP:conf/sp/WangYSLVZZ19}. A backdoor is detected if any of these triggers has a significantly smaller $\ell_1$-norm than the others and has a variance larger than a threshold, e.g., 2. The target label associated with the trigger is considered as the target label of the backdoor attack.
	
We examine the resistance of the new backdoor trigger to Neural Cleanse, where a backdoored ResNet-18 model is considered on the CIFAR-10 dataset. Table~\ref{tab:cleanse} summarizes the $\ell_1$-norm and the anomaly indices of all ten image classes, where the target label is set to be \emph{dog}. The median and the median absolute fdeviation (MAD) of the $\ell_1$-norm of the triggers are 0.000549316 and 0.00033934, respectively. We see that the anomaly index values of the first two classes (i.e., airplane and car) are greater than 2, but their $\ell_1$-norm values are greater than the median (i.e., 0.000549316). Therefore, the two classes are not considered as target classes. The remaining classes (from row 4 to row 11 in Table \ref{tab:cleanse}) all have their anomaly indices smaller than 2. As a result, the Neural Cleanse fails to detect the proposed backdoor trigger.
    
    \begin{table}
    \caption{{\color{black}Results of Neural Cleanse on the CIFAR-10 dataset and the backdoored ResNet-18 model trained with the new trigger. $m=4$. The median value is 0.000549316. The MAD is 0.00033934. The Anomaly index of the $i$-th class is $A_i = |\ell_1^{(i)} - \text{median}|/\text{MAD}$ for $i = 1,\cdots, 10$.}}
    \label{tab:cleanse}
    \centering
    \begin{tabular}{c|c|c|c|c}
    		    \hline
    			{Class}&$\ell_1$ of trigger& {$<$ Median ?} &$A_i$& {$> 2$ ?}\\
    			\hline   
    		 airplane&	0.003570557&\xmark&   8.903278025&\cmark\\
            car&	0.001647949&\xmark&	3.237555645&\cmark\\
            bird& 0.001098633&\xmark&	1.618777823&\xmark\\
            cat&	0.000732422&\xmark&	0.539592608&\xmark\\
            deer&	0.000640869&\xmark&	0.269796304&\xmark\\
            dog&	0.000457764&\cmark&	0.269796304&\xmark\\
            frog&	0.000366211&\cmark&	0.539592608&\xmark\\
            horse&	0.000366211&\cmark&	0.539592608&\xmark\\
            ship&	0.000274658&\cmark&	0.809388911&\xmark\\
            truck&	0.000274658&\cmark&	0.809388911&\xmark\\
    			\hline
    \end{tabular}
    \end{table}

\subsubsection{Spatial Smoothing}
Spatial Smoothing obfuscates a trigger's perturbation by blurring the pixel values in each patch of an image~\cite{DBLP:conf/ndss/Xu0Q18}. A window size specifies the size of a patch on the image. The window size of 1 indicates a median filter with size of $1 \times 1$.

As shown in Table~\ref{tab: defense}, the new backdoor trigger is resistant to the obfuscation by the Spatial Smoothing technique. The ASRs are close to 100\% under all the considered datasets. This is because the perturbations in each tile are identical, hence Spatial Smoothing has minimal impact on the performance of the trigger. On the other hand, Spatial Smoothing can substantially compromise the Functionality of the backdoored DNN models, particularly on high-dimensional, color image datasets. The Functionality of the backdoored DNN model drops from 87.1\% to 12.1\% on the CIFAR-10 dataset, and from 91.8\% to 2.4\% on the BSTR dataset. This is because some image features are lost as a result of the obfuscation induced by Spatial Smoothing.

\begin{table}
		\caption{The defense effect of Spatial Smoothing.}
		\label{tab: defense}
		\centering
		\begin{tabular}{c|r|r|r|r|r|r}
			\hline
			{Window} &\multicolumn{2}{c}{{MNIST}}\vline&\multicolumn{2}{c}{{CIFAR-10}}\vline&\multicolumn{2}{c}{{BTSR}}\\
			 {Size}&{Func.} &{ASR} &{Func.} &{ASR}&{Func.} &{ASR}\\
			\hline   
			{1}& 98.3\% &96.8\% &87.1\% & 95.3\% &91.8\% & 98.7\%\\
			{2}& 94.6\% &98.1\% &67.0\% & 97.5\%&35.5\%&92.9\% \\
			{3}& 97.5\% &95.2\%&22.1\% & 99.7\%&80.8\%&81.0\%\\
			{4}& 93.8\% &95.5\% &26.1\% &96.7\%&29.4\%&96.0\%\\
			{5}& 89.6\% &96.2\% &12.1\%&99.8\%&2.4\%&98.0\%\\
			\hline
		\end{tabular}
\end{table}

\subsubsection{Image Transformation}
Defenders can destroy the perturbation caused by the backdoor trigger by performing image transformations, such as random cropping, random rotation, and horizontal flipping.
Table~\ref{tab: transform_impact} demonstrates that these image transformations have no adverse effect on the ASR under the CIFAR-10 and BTSR datasets. 
The ASR is consistently around 98.6\% under the BSTR dataset, while it increases by 0.3 to 1.5\% under the CIFAR-10 dataset. The reason is twofold. Firstly, the backdoor trigger is symmetric; see Section~\ref{sec: trigger_desc}. Flipping has no impact on the performance of the trigger. Secondly, the training data contains randomly cropped and/or rotated versions of the poisoned images due to the application of data augmentation. Therefore, the DNN can still be triggered to open a backdoor by transformed images.

\begin{table}
	\caption{Impact of image transformation on Functionality (\%) and ASR (\%) }
	\label{tab: transform_impact}
	\centering
	\begin{tabular}{l|r|r|r|r|r|r}
		\hline
		&\multicolumn{2}{c}{{MNIST}}\vline&\multicolumn{2}{c}{{CIFAR-10}}\vline&\multicolumn{2}{c}{{BTSR}}\\
		&{Func.} &{ASR}&{Func.} &{ASR}&{Func.} &{ASR}\\
		\hline   
		{No Transform}& 98.3 &96.8 &87.1 & 95.3 &91.8 & 98.7\\
		{Random Crop}& 96.0 &94.0 & 86.7 & 96.8 &91.7 & 98.5\\
		{Rotation $\pm 5^{\circ}$}& 96.6 &89.2  &86.3 &96.0 &91.1 & 98.7\\
		{Horizontal Flip}& 96.6 &89.4 & 87.2 &95.6 &91.7 & 98.6\\
		\hline
	\end{tabular}
\end{table}

\subsubsection{ABS}
ABS is effective when the maximum reverse-engineered trojan trigger's ASR (RE-ASR) of a benign model is considerably lower than the RE-ASR of its trojaned version, e.g., by 5\% or more~\cite{DBLP:conf/ccs/LiuLTMAZ19}. 
Table~\ref{tab:abs_result} shows that the gap is smaller than 2\% on the MNIST and BTSR datasets. 
ABS cannot detect the backdoored MLP and ResNet-18 models trained with the new trigger. Particularly, it cannot detect suspicious neurons in the backdoored MLP model and fails to reverse-engineer any trigger. The RE-ASR is zero. 

Table~\ref{tab:abs_result} also shows that, on the CIFAR-10 dataset, the best RE-ASR is 34.16\% under the backdoored ResNet-18 model, while the maximum RE-ASR of the benign model is 24.83\%. ABS can be aware of the existence of the new backdoor trigger.
Nevertheless, the attacker can launch an adaptive attack to refine the backdoored model using the best reverse-engineered trigger generated by ABS. Specifically, we replace half of the clean images with images poisoned with the trigger in the poisoned training dataset, and continue to train the backdoored model for 30 epochs. 
The poisoned training dataset contains 5\% poisoned images tampered with the proposed trigger and the attack target class label, 47.5\% poisoned images tampered with the best reverse-engineered triggers and with their original class labels unchanged, and 47.5\% clean images with original class labels.
We see that, after the adaptive attack, the RE-ASR of the backdoored model drops to 18.26\%, even lower than the maximum RE-ASR of the benign model. ABS cannot detect the backdoored models, under adaptive attacks with the new trigger.

\begin{table}
    \centering
    \caption{{\color{black}The resistance of the new trigger to the ABS. The model trained on the MNIST dataset is an MLP. The models trained on the BTSR and CIFAR-10 datasets are ResNet-18.}}
    \renewcommand{\arraystretch}{1.2}
    \renewcommand\tabcolsep{2.5pt}
    \begin{tabular}{c|l|l|l|l|l}
        \hline
         Dataset& RE-ASR of& Max. RE-ASR& ASR& Func. of& Func. of \\
         & Trojaned &of Benign & & Benign & Trojaned \\
        \hline
       MNIST  & 0\% & 0\% &89.68\% & 99\% &96.70\%\\
        \hline
         BTSR & 16.89\%& 15.11\%&98.66\% & 93.9\%&92.95\%\\
        \hline
        CIFAR-10& \textbf{34.16\%} & 24.83\% & 91.51\%&89.5\% &90.26\%\\
        \multicolumn{1}{c}{Adaptive Attack}\vline& 18.26\% & 24.83\% & 98.88\%&89.5\% &87.98\%\\
        \hline
    \end{tabular}
    \label{tab:abs_result}
\end{table}

\subsubsection{Fine-tuning}
\begin{figure*}[!t]
     	\centering
        	\begin{subfigure}{0.425\textwidth}
    		\includegraphics[width=3.25in, height=2.25in]{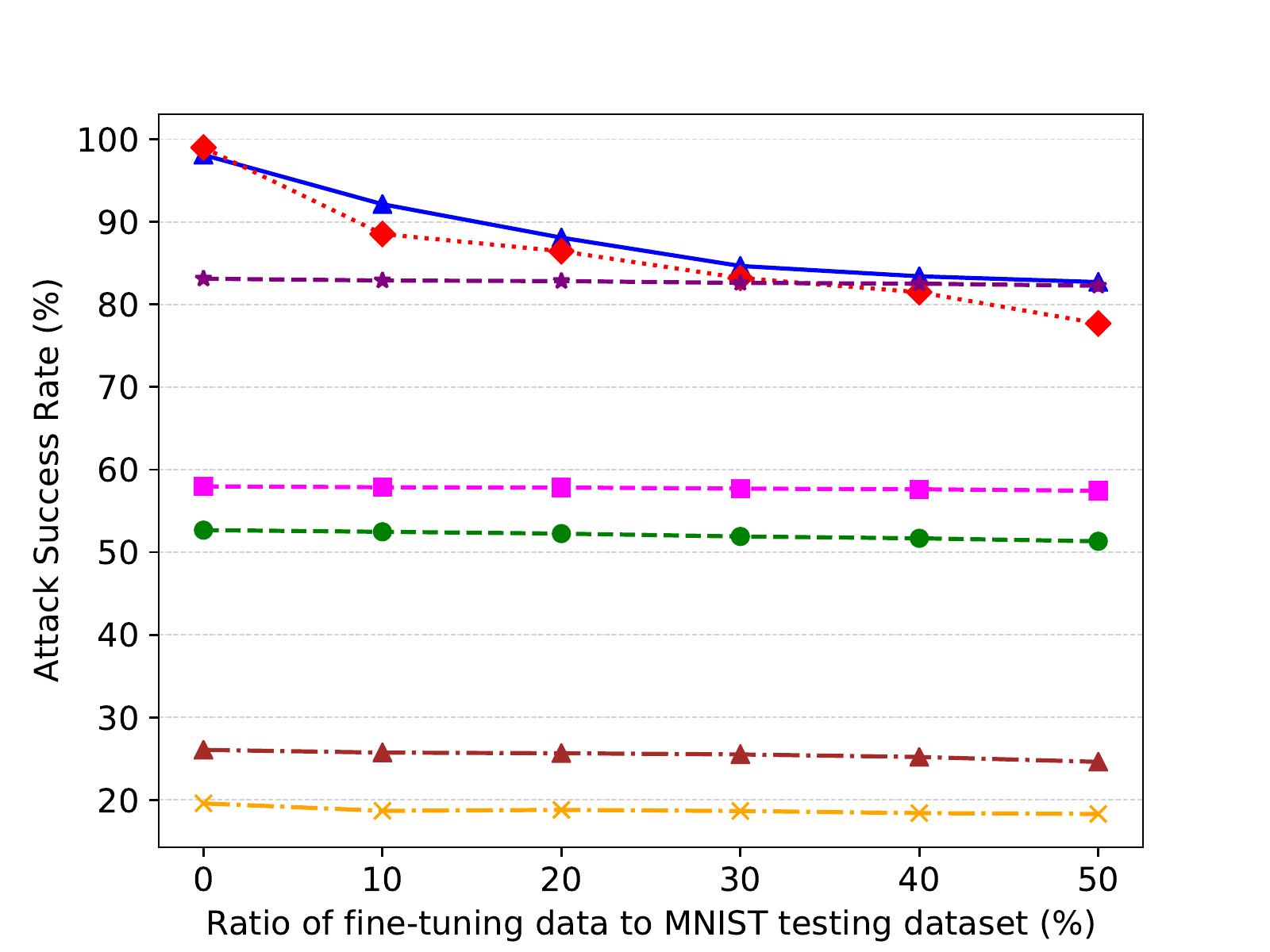}
    		\caption{ASR on the MNIST dataset}
    	\end{subfigure}
    	~~~
    	\begin{subfigure}{0.425\textwidth}
    		\includegraphics[width=3.25in, height=2.25in]{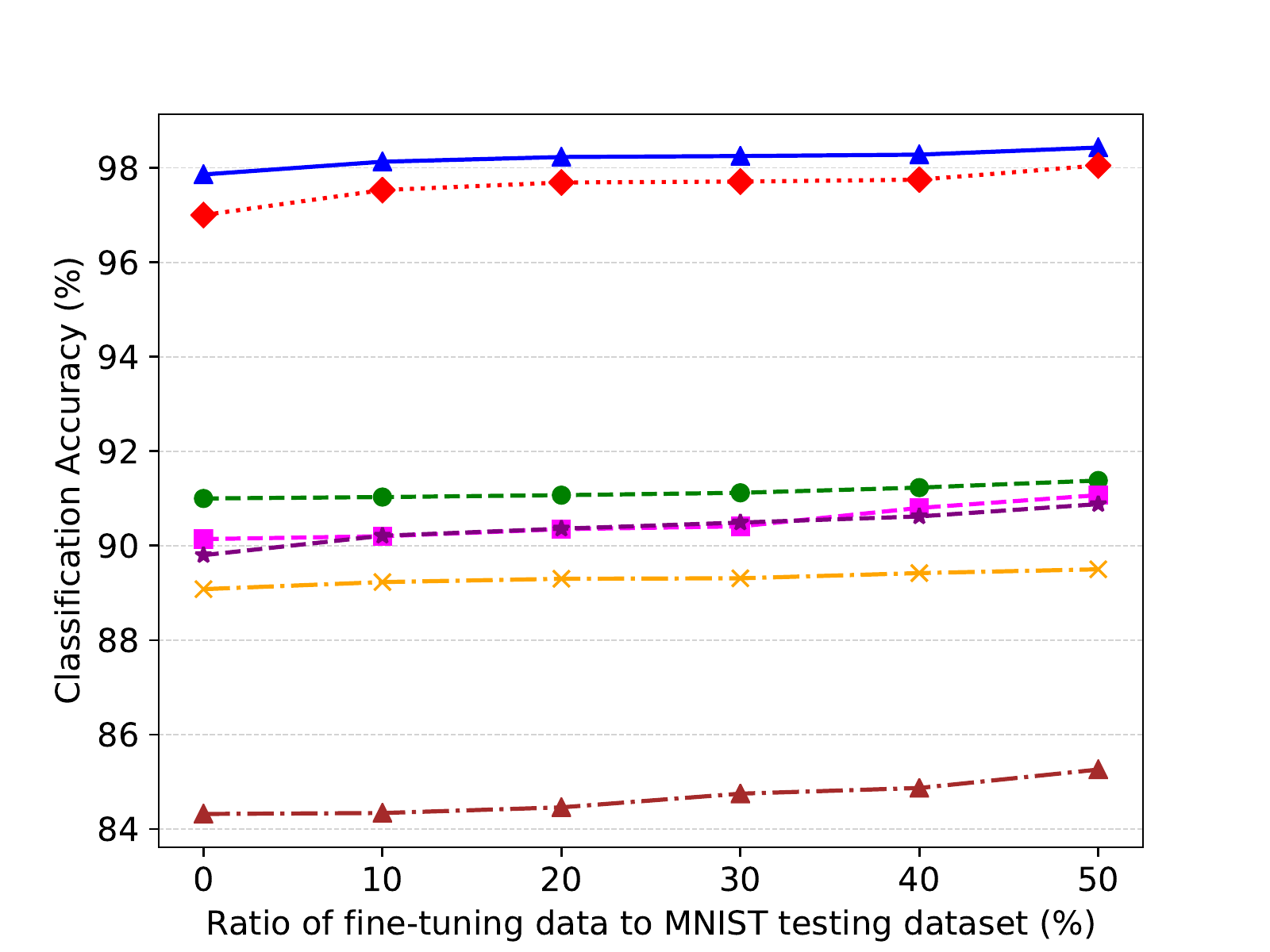}
    		\caption{Functionality on the MNIST dataset}
    	\end{subfigure}
    	
    	\begin{subfigure}{0.425\textwidth}
    		\includegraphics[width=3.25in, height=2.25in]{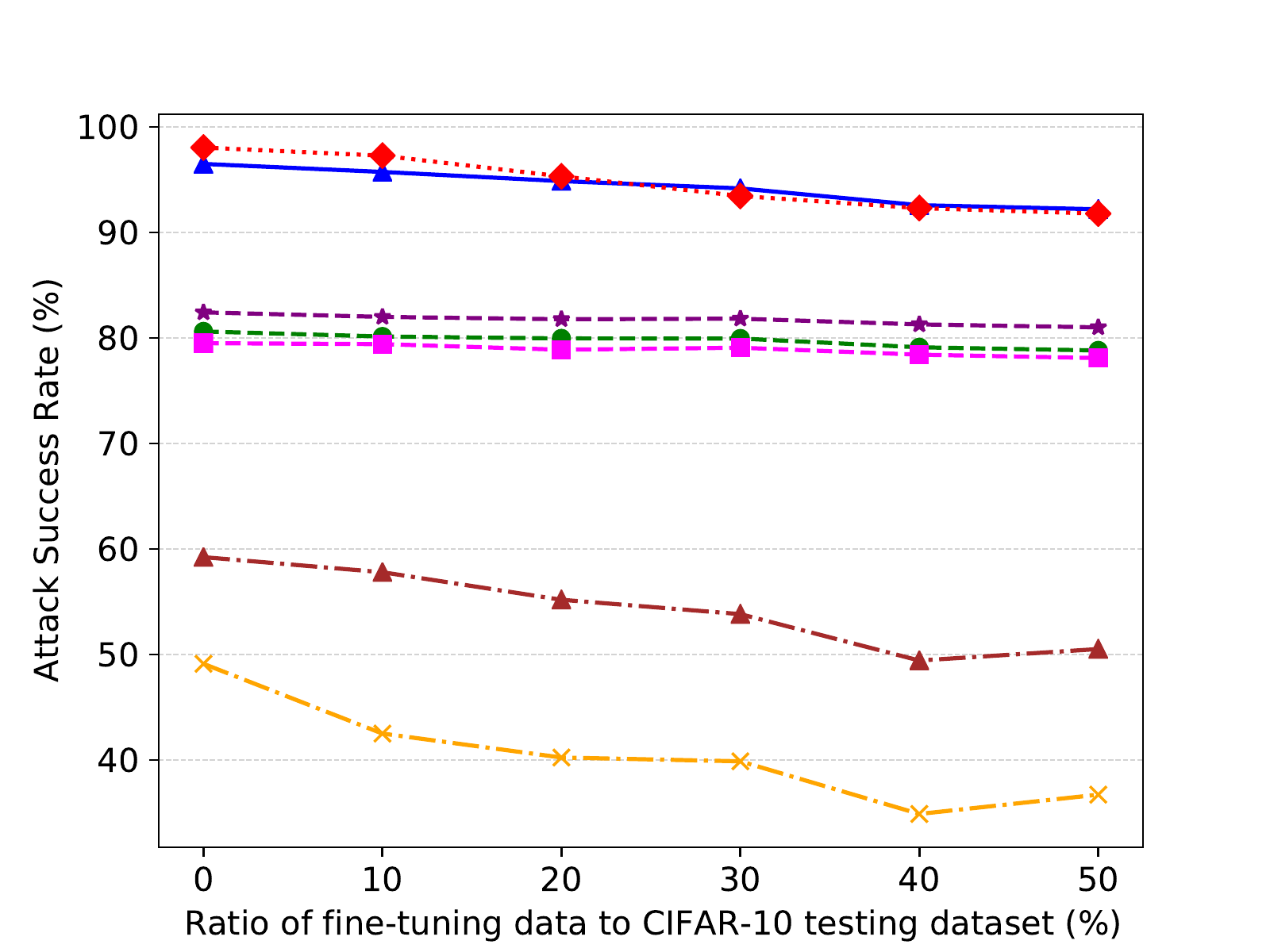}
    		\caption{ASR on the CIFAR-10 dataset}
    	\end{subfigure}
    	~~~
    	\begin{subfigure}{0.425\textwidth}
    		\includegraphics[width=3.25in, height=2.25in]{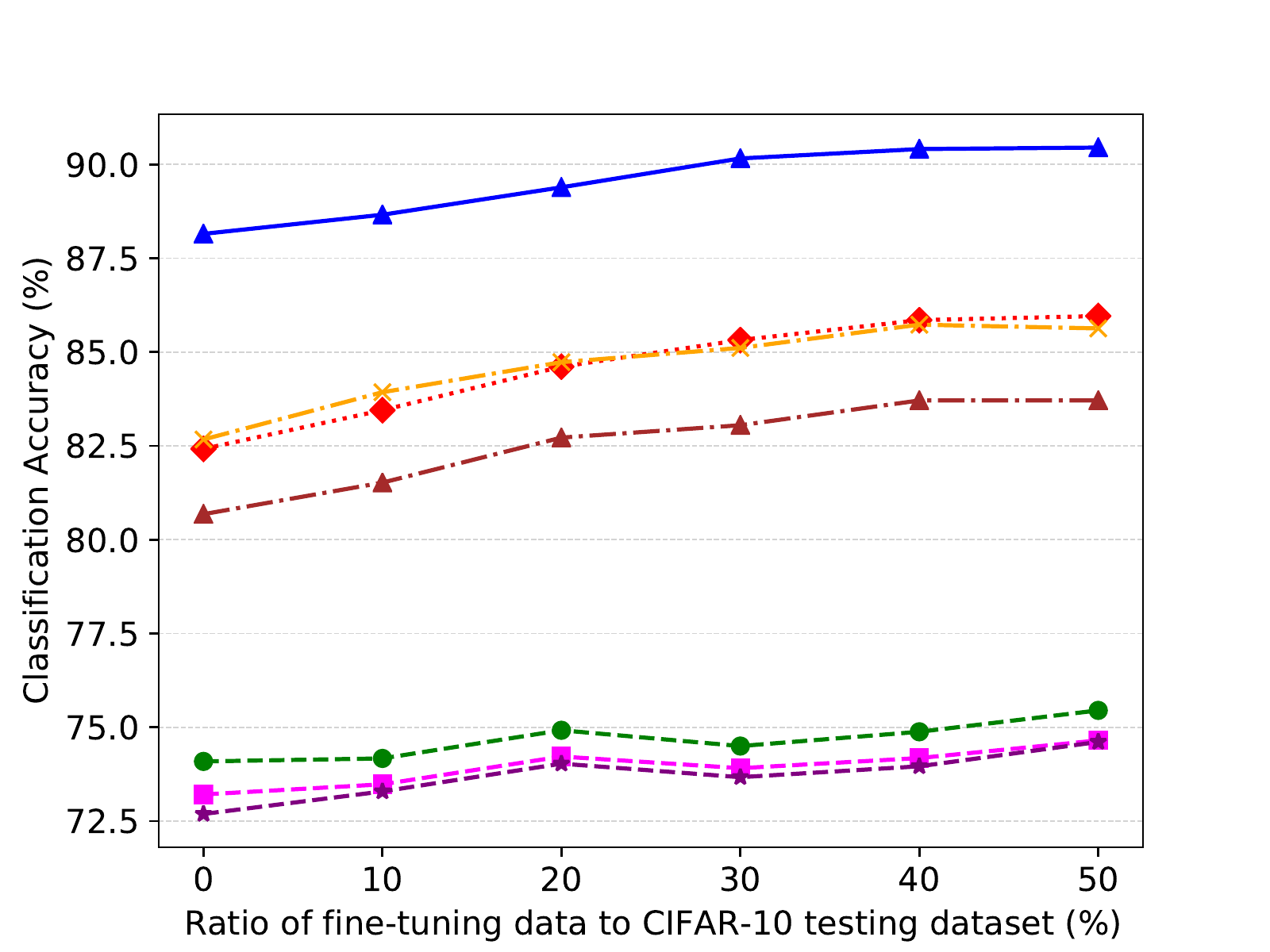}
    		\caption{Functionality on the CIFAR-10 dataset}
    	\end{subfigure}
    	
    	\begin{subfigure}{0.425\textwidth}
    		\includegraphics[width=3.25in, height=2.25in]{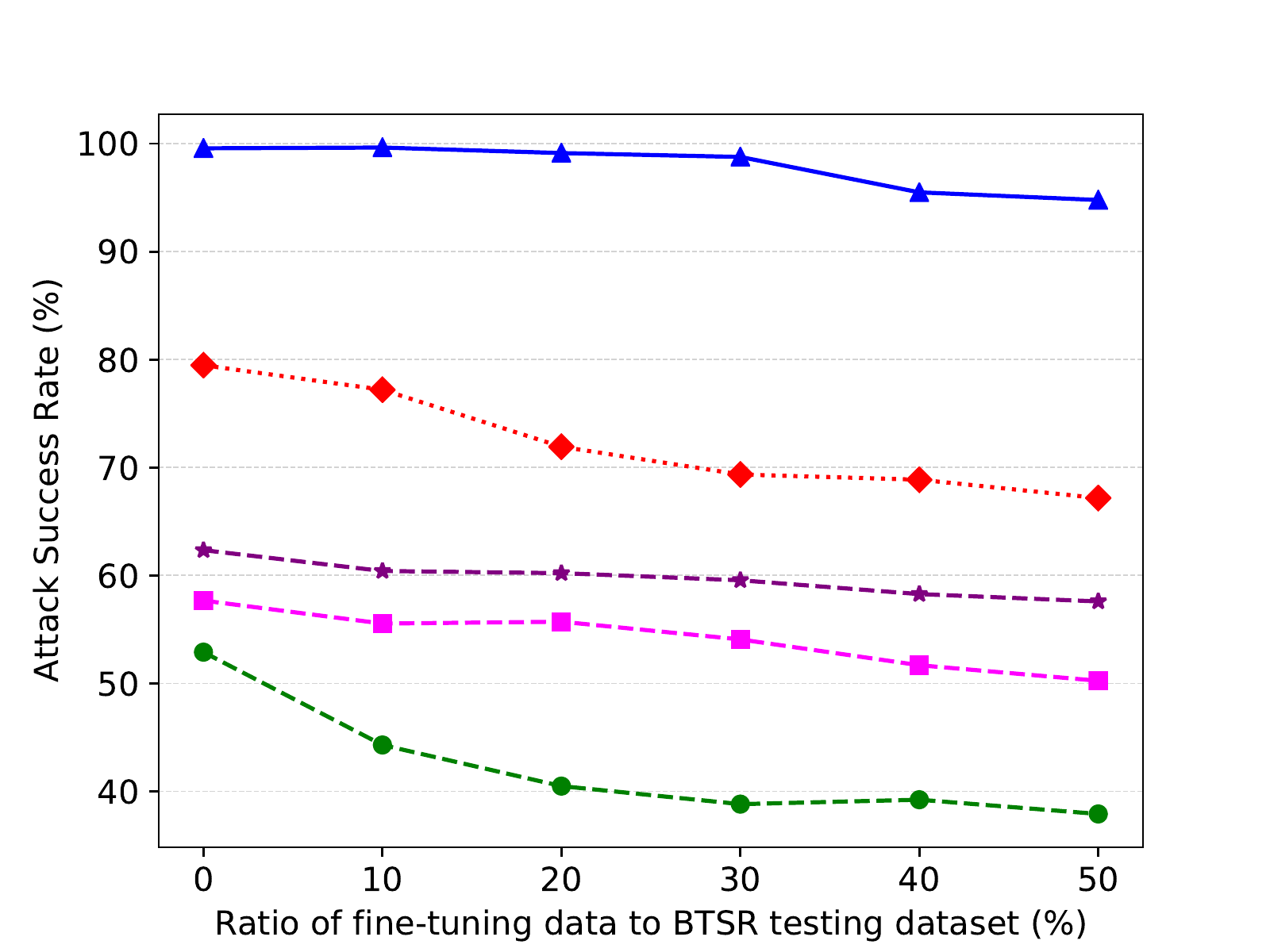}
    		\caption{ASR on the BTSR dataset}
    	\end{subfigure}
    	~~~
    	\begin{subfigure}{0.425\textwidth}
    		\includegraphics[width=3.25in, height=2.25in]{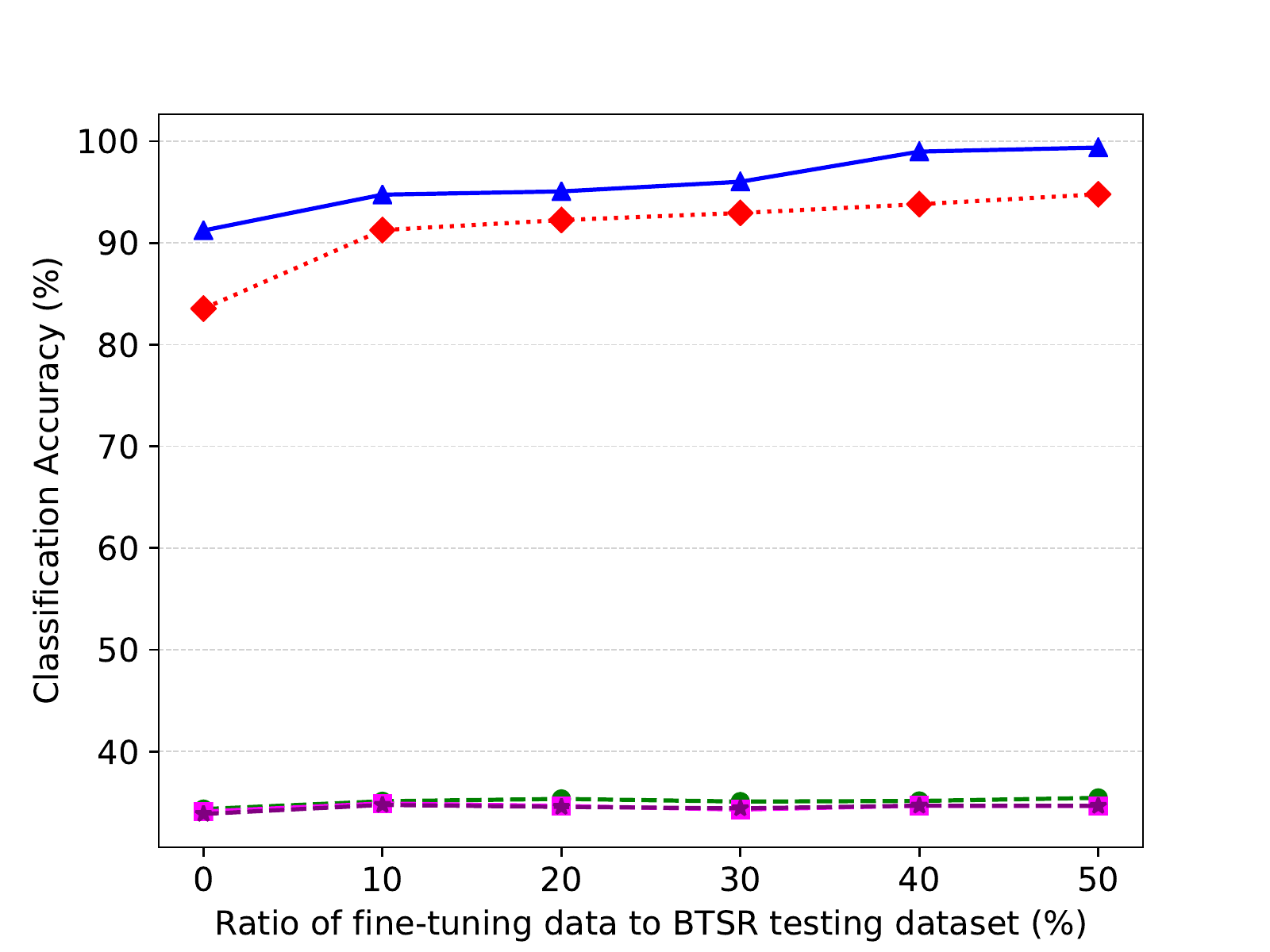}
    		\caption{Functionality on the BTSR dataset}
    	\end{subfigure}
    	
    	\begin{subfigure}{1\textwidth}
    	    \centering
    		\includegraphics[width=6in]{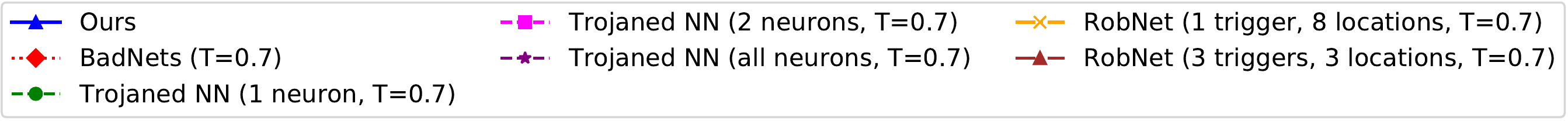}
    	\end{subfigure}
     	\caption{{\color{black}The robustness of the considered backdoor attacks against fine-tuning, where the backdoored models are fine-tuned using part of the benign testing datasets. The magnitude of our proposed trigger is  $m=10$, 6, and 5 under the MNIST, CIFAR-10, and BTSR datasets, respectively. For the clarity of the subfigures, the methods providing poor ASRs, i.e., hidden backdoor (HB) (see Figs.~\ref{fig:asr_lpips_CIFAR-10}--\ref{fig:asr_ssim_bstr}) and RobNet on the BTSR dataset (see Figs.~\ref{fig:asr_lpips_btsr}--\ref{fig:asr_ssim_bstr}), are not plotted.}} 
    	\label{fig:finetune}
    \end{figure*}

Fig.~\ref{fig:finetune} evaluates the robustness of the proposed trigger and the benchmarks against fine-tuning, where part of the testing data is repurposed by the defender (i.e., the recipient of the backdoored DNN model) to fine-tune the DNN model. The $x$-axis of the subfigures specifies the proportion of the testing data that is randomly selected to fine-tune the DNN model. The rest of the testing data is used to test the model and plot the curves. The hyperparameters of the fine-tuning are consistent with those observed at the end of the DNN model training. 
For a fair comparison, the number of epochs is 10 under all the considered methods.
	
As expected, the ASRs of the considered backdoor attacks decline and the classification accuracies of the backdoored DNN models (regarding benign inputs) improve, with the increase of benign inputs used for fine-tuning the DNN models. 
The new trigger remains the most effective after fine-tuning, offering the highest ASR and classification accuracy on all the considered datasets; see Fig.~\ref{fig:finetune}. We note that BadNets provides a similar ASR and/or classification accuracy to the proposed trigger under some of the datasets; see Figs.~\ref{fig:finetune}(a)--\ref{fig:finetune}(c). But it is considered to be less effective than the new trigger due to its much worse performance in imperceptibility (see Figs.~\ref{fig:asr_lpips_CIFAR-10} and \ref{fig:asr_ssim_CIFAR-10}) and on the other datasets; see Figs.~\ref{fig:finetune}(d)--\ref{fig:finetune}(f).

\subsection{Comparison of Run Time}
Table~\ref{tbl:complexity} provides a quantitative comparison between the proposed approach and the existing schemes in terms of the time required to generate a trigger, denoted by T-time, and the time to produce a malicious input perturbed by the trigger, denoted by M-time. 
All the experiments are carried out on a server with Intel(R) Xeon(R) Gold 6258R CPU@2.70GHz and 503G memory, and an NVIDIA A100 Tensor Core GPU with 80G memory, running Python 3.7.11, Numpy 1.21.2, and PyTorch 1.10.2 installed on an Ubuntu 18.04.5 LTS operating system.
The Python time module, time.time(), is called at the beginning and the end of a trigger or malicious input generation process to evaluate the T-time and M-time.

We can see from the third column of Table~\ref{tbl:complexity} that the trigger generation time of ours is negligible and much shorter than AdvGAN, UAT, Trojaned NN, and RobNet on all the considered datasets. We can also see from the last column of Table~\ref{tbl:complexity} that the malicious input generation time of our approach is the shortest among all of the considered methods under the MNIST dataset and the CIFAR-10 dataset. We note that the triggers used by BadNets and Hidden Backdoor (e.g., a color block) are selected in prior, and the trigger generation time (T-time) is not applicable.

\begin{table}
 \centering
 \caption{Comparison of the time required to produce a trigger (T-time) and the time required to produce a malicious input with the trigger (M-time). ``/'' indicates ``not applicable".}
 \renewcommand{\arraystretch}{1.2}
 \renewcommand\tabcolsep{2.5pt}
{\color{black}
 \begin{tabular}{c|c|c|c}
  \hline
  &Method&T-time (s) & M-time (s)\\
  \hline
  \multirow{10}{*}{\rotatebox[origin=c]{90}{MLP on MNIST }}&Ours (m=10)&$9.998\times 10^{-3}$&$8.601 \times 10^{-5}$\\
  &AdvGAN ($\epsilon =0.3$, 60 epochs)&263.55&$2.001\times 10^{-3}$\\
  &UAT ($\epsilon = 16/255$, 10 epochs)&47.21& $1.847\times 10^{-4}$\\
  &BadNets&  / &$6.893 \times 10^{-4}$\\
  &Trojaned NN (1 neuron)&4.00&$1.360 \times 10^{-4}$\\
  &Trojaned NN (2 neurons)&4.40&$1.570 \times 10^{-4}$\\
  &Trojaned NN (all neurons)&9.98&$1.374 \times 10^{-4}$\\
  &Hidden Backdoor& / &0.773\\
  &RobNet (1 trigger, 8 locations)&4.18&$1.845 \times 10^{-4}$\\
  &RobNet (3 triggers, 3 locations)&11.57&$1.310 \times 10^{-4}$\\
  \hline
  \multirow{10}{*}{\rotatebox[origin=c]{90}{ResNet-18 on CIFAR-10}}&Ours (m=6)&$1.646\times 10^{-2}$&$1.085 \times 10^{-4}$\\
  &AdvGAN ($\epsilon =8/255$, 60 epochs)&511.82&$2.050\times 10^{-3} $\\
  &UAT ($\epsilon = 10/255$, 10 epochs)&146.76&$2.191 \times 10^{-4}$\\
  &BadNets& / &$2.214 \times 10^{-4}$\\
  &Trojaned NN (1 neuron)&295.11&$2.125 \times 10^{-4}$\\
  &Trojaned NN (2 neurons)&328.70&$1.979 \times 10^{-4}$\\
  &Trojaned NN (all neurons)&366.88&$2.014 \times 10^{-4}$\\
  &Hidden Backdoor& / &0.162\\
  &RobNet (1 trigger, 8 locations)&255.08&$2.511 \times 10^{-4}$\\
  &RobNet (3 triggers, 3 locations)&1010.81&$2.014 \times 10^{-4}$\\
  \hline
  \multirow{10}{*}{\rotatebox[origin=c]{90}{ResNet-18 on BTSR}}&Ours (m=5)&$0.988$&$1.975\times 10^{-3}$\\
  &AdvGAN ($\epsilon =10/255$, 60 epochs)&485.77& $3.035\times 10^{-3}$\\
  &UAT ($\epsilon = 10/255$, 10 epochs)&29.77&$1.440 
  \times 10 ^{-3}$\\
  &BadNets& / &$1.642 \times 10^{-2}$\\
  &Trojaned NN (1 neuron)&1055.86&$0.433$\\
  &Trojaned NN (2 neurons)&1372.42&$0.402$\\
  &Trojaned NN (all neurons)&1261.80&$0.413$\\
  &Hidden Backdoor& / &8.170\\
  &RobNet (1 trigger, 8 locations)&2413.13&0.113\\
  &RobNet (3 triggers, 3 locations)&4463.91&$8.492 \times 10 ^{-2}$\\
  \hline
 \end{tabular}
 }
 \label{tbl:complexity}
\end{table}

\section{Conclusion and Future Work}
\label{sec: conclusion}
In this paper, we proposed a new backdoor trigger, which is a uniformly randomly generated 3D binary pattern and can be {\color{black}horizontally and/or vertically repeated and mirrored and} superposed onto three-channel images to train backdoored DNN models. While the new trigger collectively holds a strong recognizable pattern to effectively train or activate the backdoor of a DNN model, it generates weak perturbation to individual pixels and hence remains imperceptible. 
The complexity of the trigger generation and image perturbation is linear to the image size, and substantially lower than that of the existing triggers, such as RobNet, AdvGAN, and UAT.
Extensive experiments showed that the new trigger is more than 5, 20, and 1.5 times better than the existing backdoor attacks, such as BadNets, Trojaned NN, and Hidden Backdoor, in terms of imperceptibility (LPIPS). 
The new trigger achieves nearly 100\% ASR, and invalidates the state-of-the-art defense techniques. In the future, we will investigate countermeasures to detect and defend the new trigger.

\bibliographystyle{IEEEtran}
\bibliography{trigger}

			\begin{IEEEbiography}[{\includegraphics[width=1in,height=1.25in,clip,keepaspectratio]{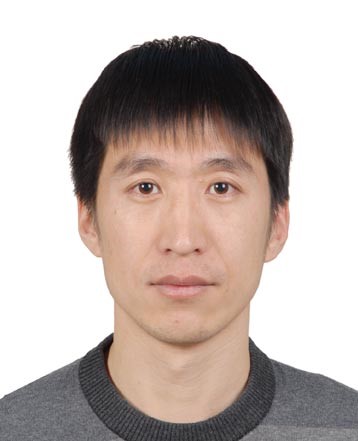}}]{Yulong Wang}
				received the Ph.D. degree in computer science and technology from Beijing University of Posts and Telecommunications (BUPT), China, in 2010. He is now an associate professor and Ph.D. supervisor with the School of Computer Science (National Pilot Software Engineering School) at BUPT. He was a visiting scientist at CSIRO, Australia from 2019 to 2020. His research interests include deep learning, software engineering, Internet-of-Things, and network security.
			\end{IEEEbiography}
			
			\begin{IEEEbiography}[{\includegraphics[width=1in,height=1.25in,clip,keepaspectratio]{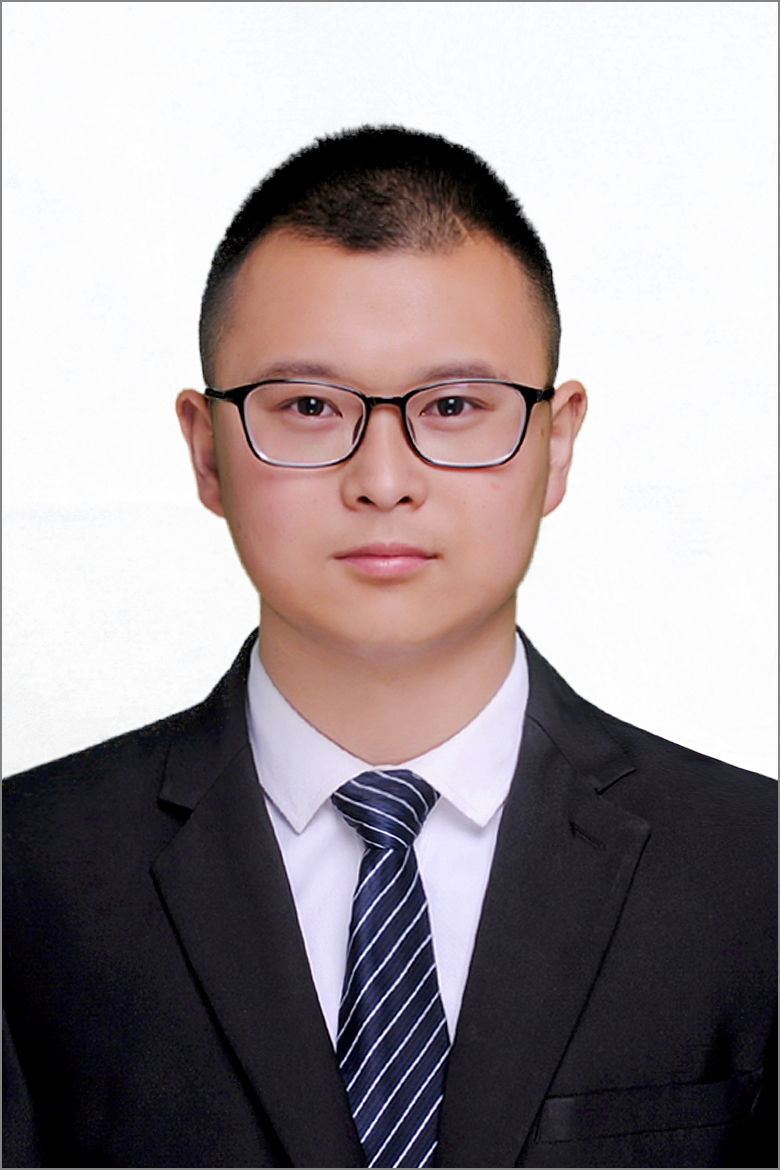}}]{Minghui Zhao}
			 	received the B.E. degree in Computer Science and Technology from Xidian University, Xi'an, China, in 2019. He received the master's degree in Computer Science and Technology, Beijing University of Posts and Telecommunications, Beijing, China, in 2022. His research interests include deep learning, software engineering, and network security.
			\end{IEEEbiography}
			
			\begin{IEEEbiography}[{\includegraphics[width=1in,height=1.25in,clip,keepaspectratio]{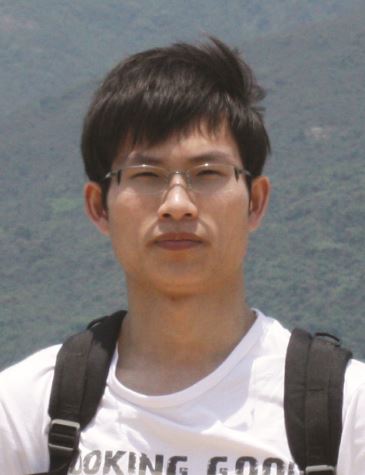}}]{Shenghong Li}
				received the B.S. degree in communication engineering from Nanjing University,
				Nanjing, Jiangsu, China, in 2008, and the Ph.D. degree in electronic and computer engineering from
				Hong Kong University of Science and Technology (HKUST), Hong Kong, in 2014, respectively. He joined CSIRO as an OCE postdoctoral fellow in 2014 and has been a Research Scientist with the Communications and Signal Processing team since 2017. His research interests include wireless tracking, cooperative localization, data fusion in localization systems, and wireless communication.
			\end{IEEEbiography}

		\begin{IEEEbiography}[{\includegraphics[width=1in,height=1.25in,clip,keepaspectratio]{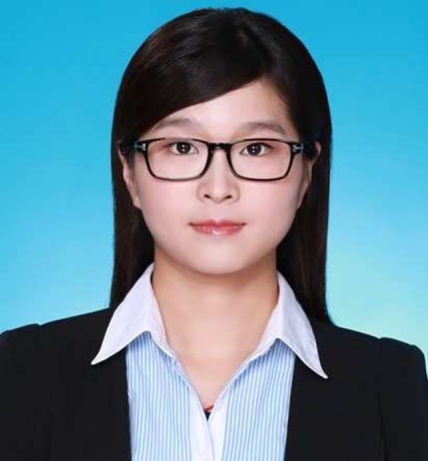}}]{Xin Yuan} received the B.E. degree from Taiyuan University of Technology, Shanxi, China, in 2013, and the dual Ph.D. degree from Beijing University of Posts and Telecommunications (BUPT), Beijing, China, and the University of Technology Sydney (UTS), Sydney, Australia, in 2019 and 2020, respectively. She is currently a Research Scientist at CSIRO, Sydney, NSW, Australia. Her research interests include machine learning and optimization, and their applications to Internet-of-Things and intelligent systems.
		\end{IEEEbiography}

\begin{IEEEbiography}
[{\includegraphics[width=1in,height=1.25in,clip,keepaspectratio]{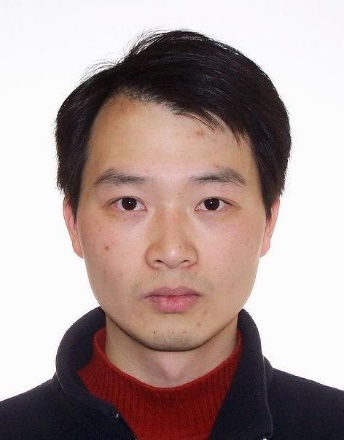}}]{Wei Ni}
(M'09-SM'15) received the B.E. and Ph.D. degrees in Electronic Engineering from Fudan University, Shanghai, China, in 2000 and 2005, respectively. Currently, he is a Principal Research Scientist at CSIRO, Sydney, Australia, an Adjunct Professor at the University of Technology Sydney, and an Honorary Professor at Macquarie University. He was a Postdoctoral Research Fellow at Shanghai Jiaotong University from 2005 to 2008; Deputy Project Manager at the Bell Labs, Alcatel/Alcatel-Lucent from 2005 to 2008; and Senior Researcher at Devices R\&D, Nokia from 2008 to 2009. He has authored five book chapters, more than 200 journal papers, 100 conference papers, 25 patents, and ten standard proposals accepted by IEEE. His research interests include machine learning, online learning, stochastic optimization, and their applications to system efficiency and integrity.          

Dr Ni is the Chair of IEEE Vehicular Technology Society (VTS) New South Wales (NSW) Chapter since 2020, an Editor of IEEE Transactions on Wireless Communications since 2018, and an Editor of IEEE Transactions on Vehicular Technology. He served first as the Secretary and then the Vice-Chair of IEEE NSW VTS Chapter from 2015 to 2019, Track Chair for VTC-Spring 2017, Track Co-chair for IEEE VTC-Spring 2016, Publication Chair for BodyNet 2015, and Student Travel Grant Chair for WPMC 2014. 
\end{IEEEbiography}

\end{document}